**Chapter 12**

# An Overview about Emerging Technologies of Autonomous Driving

Yu Huang[*], Yue Chen[**] and Zijiang Yang[***]

*Roboraction.AI, Sunnyvale, USA,

**Futurewei Technology Inc., Santa Clara, USA

***Synkrotron Technologies Co Ltd., Hefei, China

yu.huang07@gmail.com

Since DARPA's Grand Challenges (rural) in 2004/05 and Urban Challenges in 2007, autonomous driving has been the most active field of AI applications. This paper gives an overview about technical aspects of autonomous driving technologies and open problems. We investigate the major fields of self-driving systems, such as perception, mapping & localization, prediction, planning & control, simulation, V2X and safety etc. Especially we elaborate on all these issues in a framework of data closed loop, a popular platform to solve the "long tailed" autonomous driving problems.

## 1. Introduction

Autonomous Driving has been active for more than 10 years. In 2004 and 2005, DARPA held the Grand Challenges in rural driving of driverless vehicles. In 2007, DAPRA also held the Urban Challenges for autonomous driving in street environments. Then professor S. Thrun at Stanford university, the first-place winner in 2005 and the second-place winner in 2007, joined Google and built Google X and the self-driving team.

Recently there have been three survey papers about self-driving [3,9,14]. Autonomous driving, as one of the most challenging applications of AI with machine learning and computer vision etc., actually has been shown to be a "long tailed" problem, i.e. the corner cases or safety-critical scenarios occur scarcely. In this paper, we investigate how autonomous driving has been going on research and development in a data closed loop. Our overview work spans the state-of-art technology in major fields of self-driving technologies, such as perception, mapping and localization, prediction, planning and control, simulation, V2X and safety etc.

In conclusion, we will also mention the influence of the emergent Foundation Models to autonomous driving community.

## 2. Brief Introduction of Autonomous Driving

There have been some survey papers about the self-driving technologies, from the whole system/platform to individual modules/functions [1-2, 4-8, 10-13, 15-33]. In this section, we briefly introduce the basic autonomous driving functions and modules, shown in Fig. 1, hardware and software architecture, perception, prediction, mapping and localization, planning, control, safety, simulation, and V2X etc.

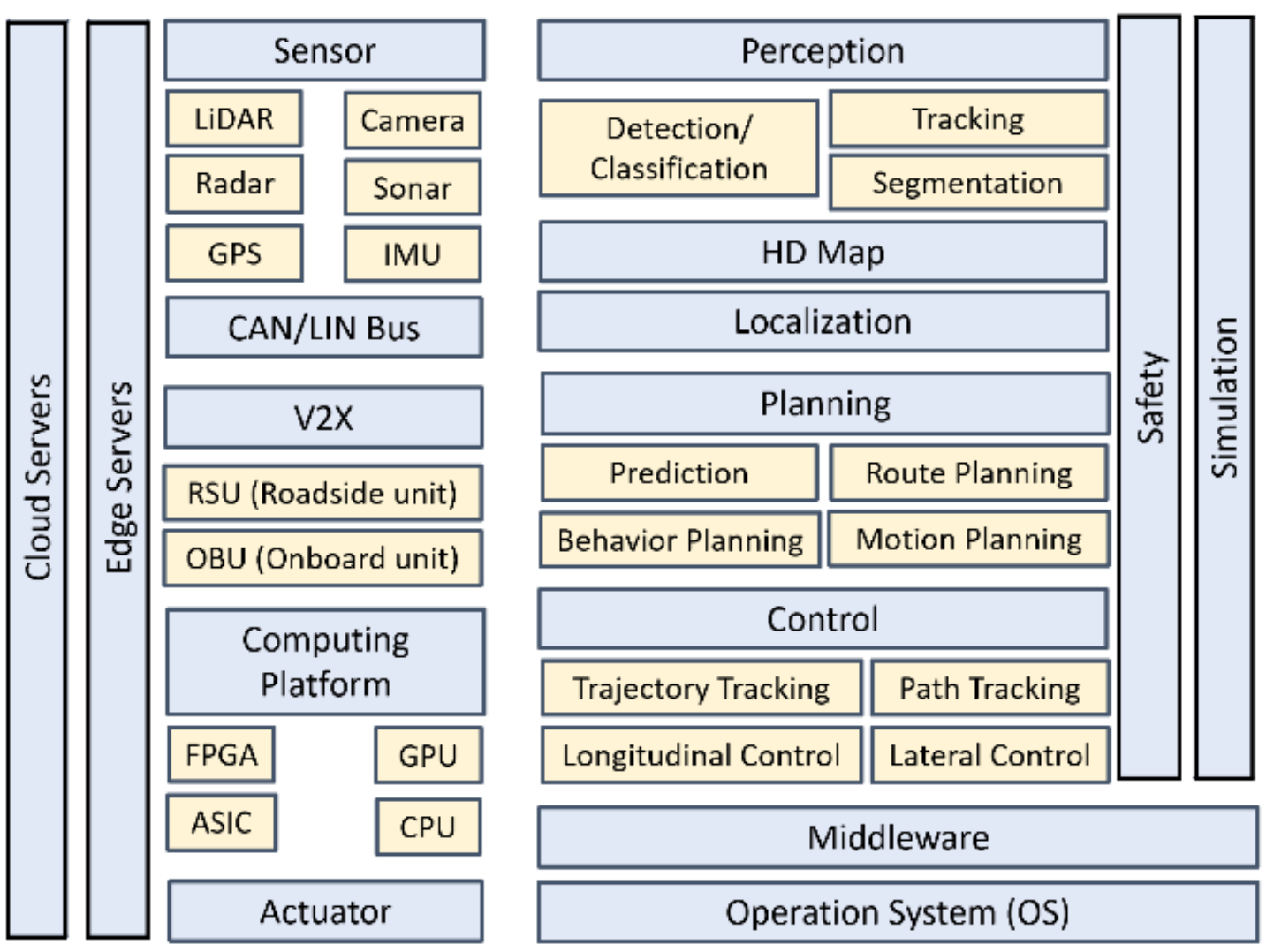

Fig. 1 HW and SW of the autonomous driving platform.

### 2.1. *Automation Levels*

The US Department of Transportation and the *National Highway Traffic Safety Administration* (NHTSA) had adopted the *Society of Automotive Engineers* (SAE) international standard for automation levels which define autonomous vehicles from Level 0 (the human driver has full control) to Level 5 (the vehicle completely drives itself).

In level 1, the driver and the automated system control the vehicle together. In level 2 the automated system takes full control of the vehicle, but the driver must be prepared to intervene immediately at any time. In level 3, the driver can be free from the driving tasks and the vehicle will call for an immediate response, so the driver must still be prepared to intervene within some limited time. In level 4, it is the same to level 3, but no driver attention is ever required for safety, e.g. the driver may safely go to sleep or leave the driver's seat.

### 2.2. *Hardware*

Autonomous driving vehicle test platforms should be capable of realizing real-time communication, such as in *controller area network* (CAN) buses and Ethernet, and can accurately complete and control the directions, throttles, and brakes of vehicles in real time. Vehicle sensor configurations are conducted to meet the reliability requirements of environmental perception and to minimize production cost.

Sensing of autonomous driving vehicles falls into three main categories: *self-sensing, localization and surrounding sensing*. Self-sensing measures the current vehicle state, i.e. velocity, acceleration, yaw, and steering angle etc. with *proprioceptive sensors*. *Proprioceptive sensors* include odometers, *inertial measurement units* (IMUs), gyroscopes, and the CAN bus. *Localization*, using external sensors such as *global positioning system* (GPS) or dead reckoning by IMU readings, determines the vehicle' s global and local position. *Surrounding sensing* uses *exteroceptive sensors* to perceive road markings, road slope, traffic signs, weather conditions and obstacles.

Proprioceptive and exteroceptive sensors can be categorized as either active or passive sensors. *Active* sensors emit energy in the form of electromagnetic waves and measure the return time to determine parameters such as distance. Examples include sonar, radar, and *Light Detection And Ranging* (LiDAR) sensors. *Passive* sensors do not emit signals, but rather perceive electromagnetic waves already in the environment (e.g., light-based and infrared cameras).

Another important issue is the computing platform, which supports sensor data processing to recognize the environments and make the real-time control of the vehicles through those computationally intensive algorithms of optimization, computer vision and machine learning. There are different computing platforms, from CPUs, GPUs, ASIC to FPGAs etc. To support AI-based autonomous driving, cloud servers are required to support big data service, such as large-scale machine learning and large size data storage (for example, HD Map). To support vehicle-road collaboration, edge communication and computing devices are required from both the vehicle side and the roadside. An example of Sensor configuration (from NuScene) in a self-driving car is shown in Fig. 2. It is installed with LiDARs, cameras, radars, GPS and IMU etc.

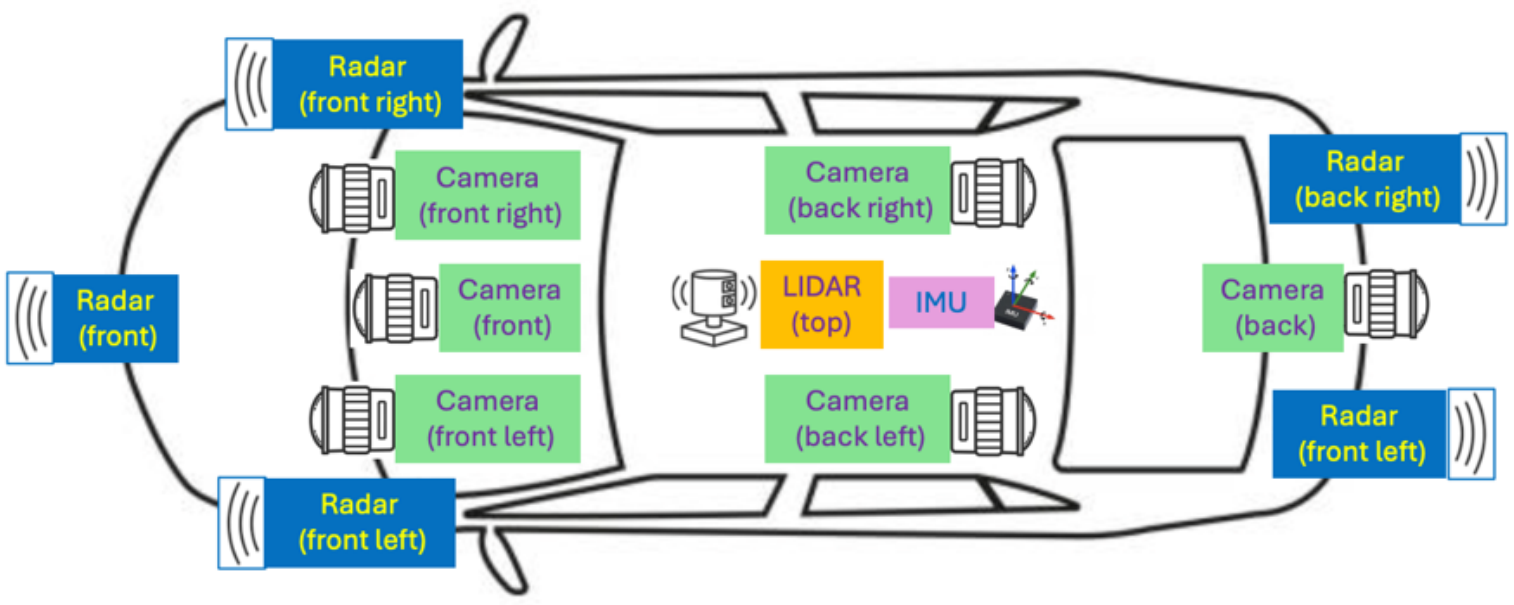

Fig. 2 An example of Autonomous Driving Sensor Hardware

If multi-modal sensor data needs to be collected, sensor calibration is required, which involves determining the coordinate system relationships between each sensor data, such as camera calibration, camera - LiDAR calibration, LiDAR - IMU calibration, and camera -radar calibration. In addition, a unified clock needs to be used between sensors (GNSS as an example), and then a certain signal is used to trigger the operation of the sensor. For example, the transmission signal of the lidar can trigger the exposure time of the camera, which is time synchronized.

### 2.3. *Software*

A software platform of autonomous driving is classified multiple layers, from bottom to top as the *real time operating system* (RTOS), middleware, function software and application software. The software architecture could be end-to-end or modular style.

*End-to-end* systems generate control signals directly from sensory inputs. Control signals can be operation of steering wheel and pedals (throttles and brakes) for acceleration/deceleration (even stop) and turn left/right. There are three main approaches for end-to-end driving: direct supervised deep learning, neuro evolution and deep reinforcement learning.

*Modular* systems are built as a pipeline of multiple components connecting sensory inputs to actuator outputs. Key functions of a modular autonomous driving system (ADS) are regularly summarized as: perception, localization and mapping, prediction, planning and decision making, and vehicle control etc. Fig. 3 illustrates E2E and modular system.

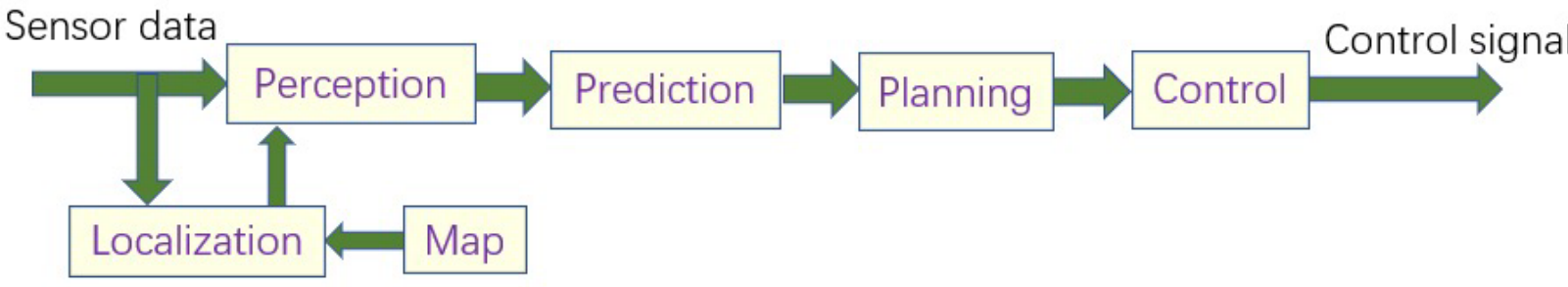

(a) Modular system

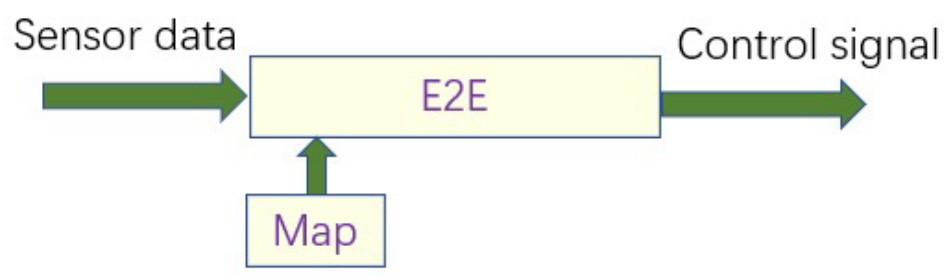

(b) End-to-end system

Fig. 3 Architecture of Autonomous Driving Software

•*Perception* collects information from sensors and discovers relevant knowledge from the environment. It develops a contextual understanding of driving environment, such as detection, tracking and segmentation of obstacles, road signs/marking and free space drivable areas. Based on the sensors implemented, the environment perception task can be tackled by using LIDARs, cameras, radars or a fusion between these three kinds of devices. At the highest level, perception methods can fall into three categories: *mediated perception, behavior reflex perception, and direct perception*. Mediated perception develops detailed maps of the surroundings as vehicles, pedestrians, trees, road markings, etc. Behavior reflex perception maps sensor data (image, point cloud, GPS location) directly to driving maneuvers. Direct perception combines behavior reflex perception with the metric collection of the mediated perception approach.

•*Mapping* refers to building the map with information of roads, lanes, signs/markings and traffic rules etc. In general, there are two main categories of maps: planar which refers to maps that rely on layers or planes on a Geographic Information System (GIS), e.g. High Definition (HD) maps, and point-cloud which refers to maps based on a set of data points in the GIS. The HD map contains some useful static elements like lanes, buildings, traffic lights, and road markings etc., necessary for autonomous driving, even for the object that cannot be appropriately detected by sensors due to occlusion. HD maps for autonomous driving tightly connects with vehicle localization functionality and keeps interacting with vehicle sensors, such as lidar, radar, and camera, to construct the perception module of the autonomous system.

•*Localization* determines its position with respect to the driving environment. *Global Navigation Satellite Systems* (GNSS) such as GPS, GLONASS, BeiDou, and Galileo rely on at least four satellites to estimate global position at a relatively low cost. GPS accuracy can be improved upon by using Differential GPS. GPS is often integrated with IMU to design a low-cost vehicle localisation system. IMUs have been used to estimate vehicle position relative to its initial position, in a method known as Dead Reckoning. Since the HD map has been used for self-driving, localization based on that is also taken into account. Recently, many studies have emerged on self-contained odometry methods and *simultaneous localization and mapping* (SLAM). Usually the SLAM techniques apply an odometry algorithm to obtain the pose where later fed into a global map optimization algorithm. Visual SLAM is still a challenging direction due to drawbacks of image-based computer vision algorithms, like feature extraction and matching, camera motion estimation, 3-D reconstruction (triangulation) and optimization (bundle adjustment).

•*Prediction* refers to estimating the obstacles' trajectories based on their kinematics, behaviors and long-term/short-term histories. To fully solve the trajectory-prediction problem, social intelligence is very important, since diverse possibilities has to be bound, the infinite search space must be limited, given the known social intelligence. To model the social interaction, we need to understand the dynamics of the agents and their surroundings to predict their future behavior and prevent any crashes.

•*Planning* makes decisions on taking the vehicle to the destination while avoiding obstacles, which generates a reference path or trajectory. Planning can be classified as route (mission) planning, behavior planning and motion planning at different levels.

- *Route planning* is referred as finding the point-to-point shortest path in a directed graph, and conventional methods are examined under four categories as goal-directed, separator-based, hierarchical and bounded-hop techniques.
- *Behavioral planning* decides on a local driving task that progresses the vehicle towards the destination and abides by traffic rules, traditionally defined by a finite state machine (FSM). Recently imitation learning and reinforcement learning are being researched to generate the behavior taken for vehicle navigation.
- *Motion planning* then picks up a continuous path through the environment to accomplish a local driving task, for example RRT (rapidly exploring random tree) and Lattice planning.

•*Control* executes the planned actions by selecting appropriate actuator inputs. Usually control could be split into *lateral* and *longitudinal* control. Mostly the control design is decoupled into two stages, trajectory/path generation and tracking, for example the pure pursuit method. However, it can generate the trajectory/path and track both simultaneously.

•*V2X (vehicle to everything)* is a vehicular technology system that enables vehicles to communicate with the traffic and the environment around them, including vehicle-to-vehicle communication (V2V) and vehicle-to-infrastructure (V2I). From mobile devices of pedestrians to stationary sensors on a traffic light, an enormous amount of data can be accessed by the vehicle with V2X. By accumulating detailed information from other peers, drawbacks of the ego vehicle such as sensing range, blind spots and insufficient planning may be alleviated. The V2X helps in increasing safety and traffic efficiency. How to collaborate between either vehicle-vehicle or vehicle-road, is still challenging.

•It is worth to mention, the *ISO* (International Organization for Standardization) *26262* standard for *functional safety* of driving vehicles defines a comprehensive set of requirements for assuring safety in vehicle software development. It recommends the use of a Hazard Analysis and Risk Assessment (*HARA*) method to recognize hazardous events and to define safety goals that mitigate the hazards. Automotive Safety Integrity Level (*ASIL*) is a risk classification scheme defined in ISO 26262 in an automotive system. AI system brings more problems in safety, which are handled by a new built standard, called ISO/PAS 21448 SOTIF (safety of the intended functionality).

•Besides of either modular or end-to-end system, there is an important platform "*simulation*" in ADS development. Since the driving of an experimental vehicle on the road still costs highly and experiments on existing human driving road networks are restricted, a simulation environment is beneficial for developing certain algorithms/modules before real road tests. A *simulation* system consists of the following core components: sensor modelling (cameras, radar, LiDAR and sonar), vehicle dynamics and kinematic, shape and kinematic modelling of pedestrians, motorists and cyclists, road network and traffic network, 3-D virtual environment (urban and rural scenes) and driving behavior modelling (age, culture, race etc.). The critical problems existing in simulation platform are "sim2real" and "real2sim", where the former refers as how to simulate realistic scenarios, and the latter as how to make scenario reproduction as digital twins.

## 3. Perception

Perceiving the surrounding environment and extracting information is the critical task for self-driving. A variety of tasks, using different sensing modalities, fall under the category of perception [5-6, 25, 29, 32, 36]. Cameras are the most commonly used sensor based on computer vision techniques, with 3D vision becoming a strong alternative/supplement.

Recently BEV (bird's eye view) perception [25, 29] has been the most active perception direction in autonomous driving, especially in vision-based systems, for two major benefits. First, BEV representations of the driving scene can be directly deployed by downstream driving applications such as trajectory prediction and motion planning etc. Second, BEV provides a interpretable way to fuse information from different views, modalities, time series, and agents. For example, other popularly used sensors, like LiDAR and Radar, which capture data in 3D space, can been easily transformed to BEV, and conduct sensor fusion with cameras directly.

In the survey paper [25], the BEV work can be categorized as below, shown in Fig. 4.

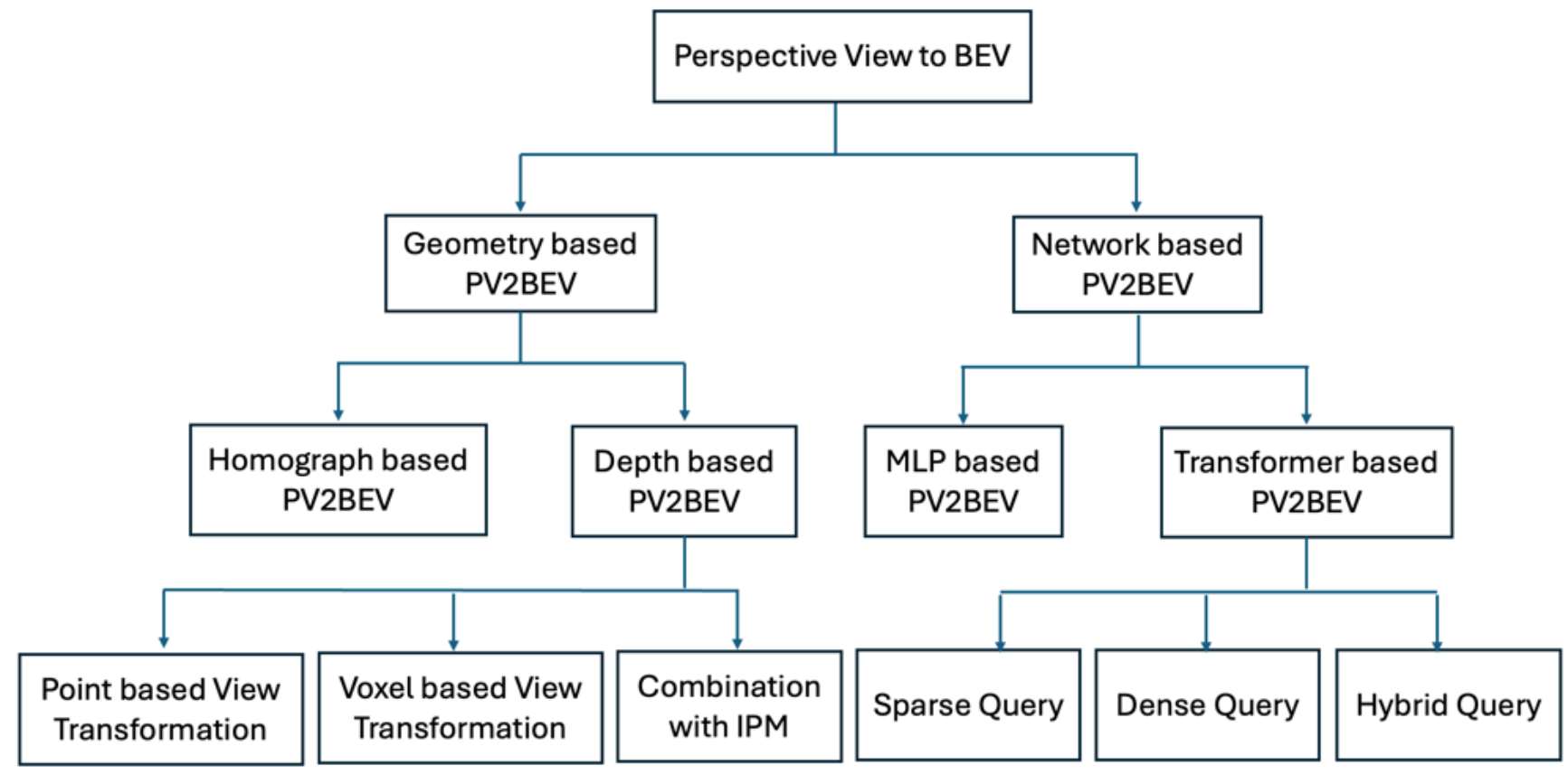

Fig. 4 Categories of BEV algorithms [25]

First, based on view transformation, it is classified as *geometry-based transformation* and *network-based transformation*; The geometry-based ones fully utilizes the physical principles of the camera to transfer the view, which can be further split into classical *homograph-based methods* (i.e. inverse projective mapping) and *depth-based methods* lifting 2D features to 3D space by explicit or implicit depth estimation.

Based on the way to utilize depth information, we can separate the depth-based methods into two types: point-based and voxel-based; the point-based methods directly use the depth estimation to convert pixels to point clouds, scattering in continuous 3D space; the voxel-based methods instead typically scatter 2D features (instead of points) at the corresponding 3D locations directly with the depth guidance.

The network-based methods may employ a bottom-up strategy where neural networks act like the view projector, another option could employ a top-down strategy by directly constructing BEV queries and searching corresponding features on front-view images by the cross attention mechanism (Transformer-based), in which sparse, dense, or hybrid queries are proposed to match different downstream tasks.

So far, the BEV network has been used for object detection, semantic segmentation, online mapping, sensor fusion and trajectory prediction etc. As illustrated in Fig. 5 from the survey paper [29], there are two typical pipeline designs for BEV fusion algorithms. The main difference lies in 2D to 3D conversion and fusion modules. In the perspective view pipeline (a), results of different algorithm are first transformed into 3D space, then fused using prior or hand-craft rules. The BEV pipeline (b) first transforms perspective view features to BEV, then fuses features to obtain the final predictions, thereby maintaining most original information and avoiding hand-crafted design.

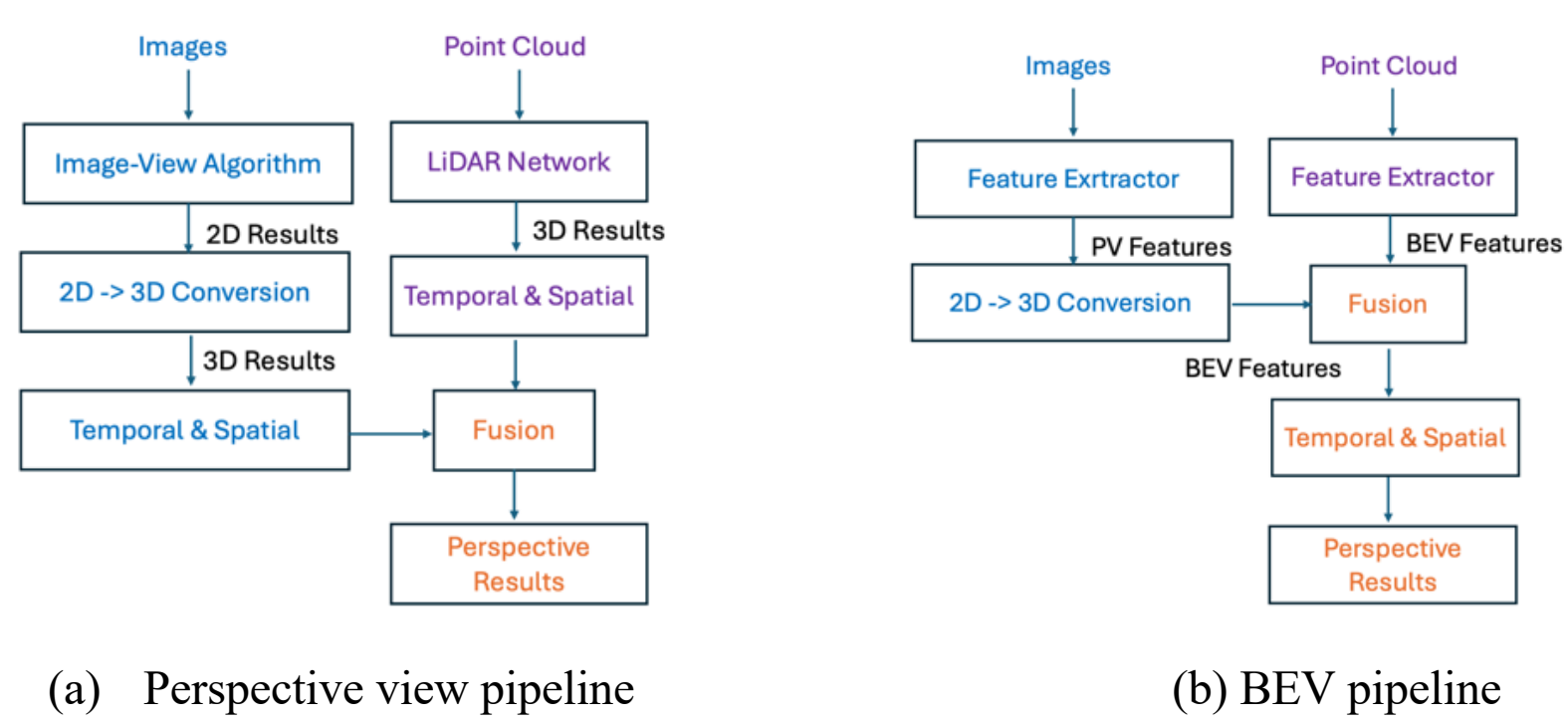

(a) Perspective view pipeline (b) BEV pipeline

Fig. 5 BEV fusion pipeline (camera and LiDAR) [29]

After BEV, 3-D occupancy network is coming to the forefront in the perception domain of self-driving field [32]. While BEV can simplify the vertical geometry of driving scenes, 3D voxels are able to represent the full geometry with a rather low resolution, including the road ground and the volume of obstacles, which pay higher computation costs. Camera-based methods are emerging in 3D occupancy network. Images are naturally dense in pixels but need depth information to be reversely projected to 3D occupancy. Note: for LiDAR data, the occupancy network actually realizes a *semantic scene completion* (SSC) task.

In Fig. 6, we give three examples of both BEV and occupancy network for camera input only, LiDAR input only and both camera and LiDAR input.

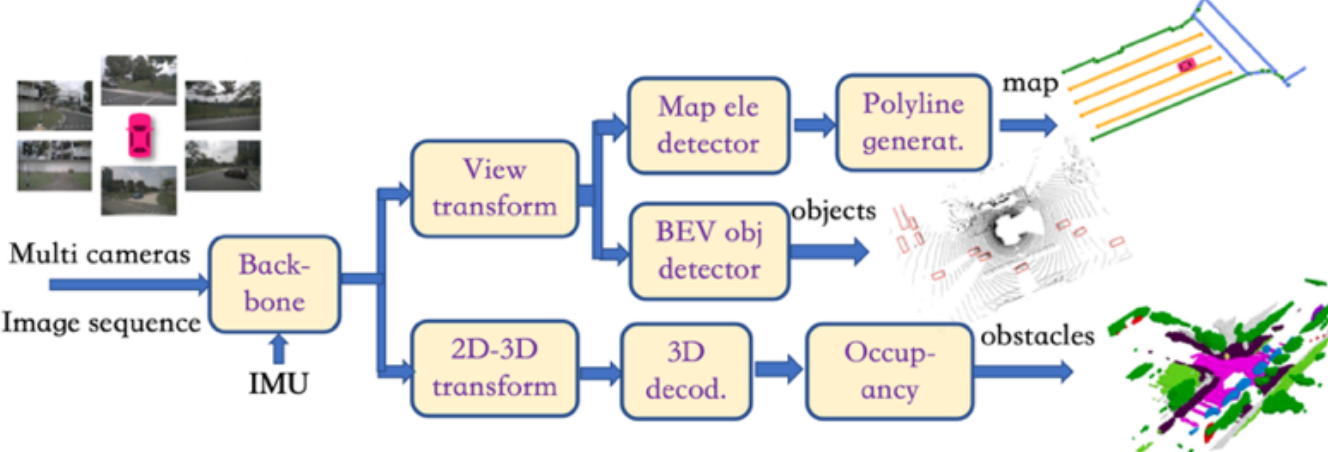

(a) camera only

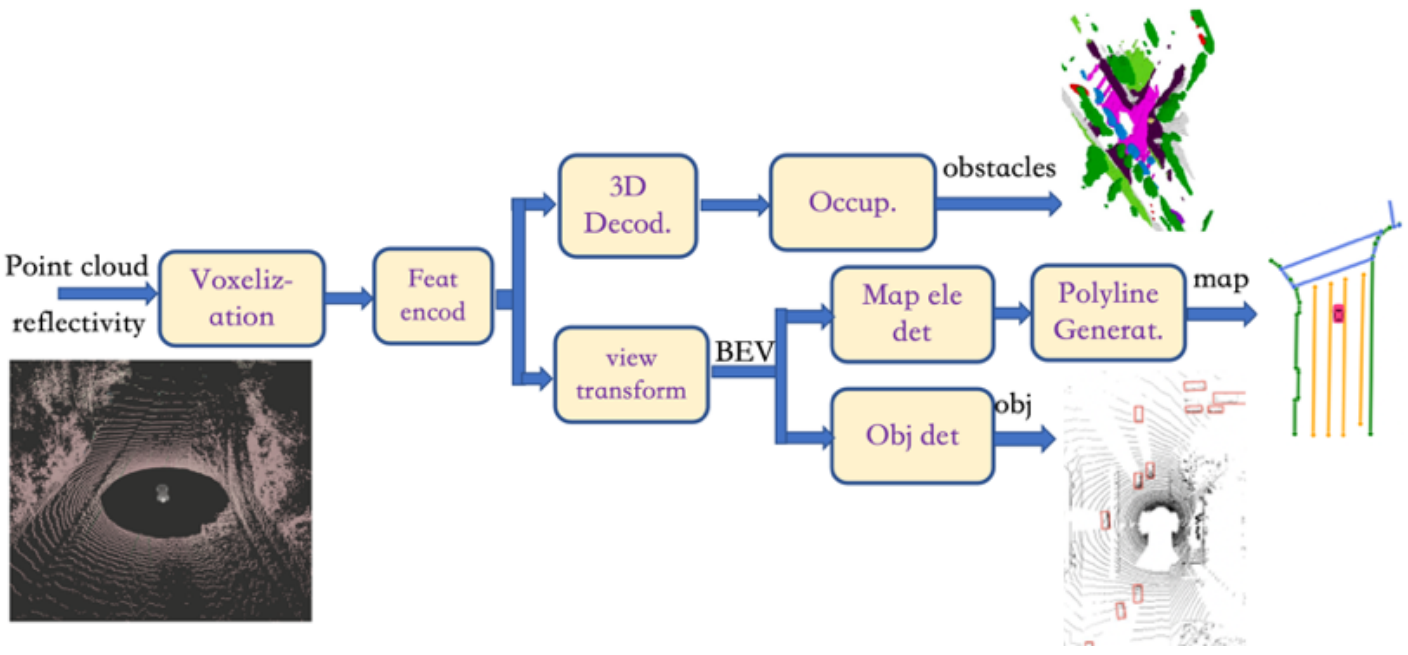

(b) LiDAR only

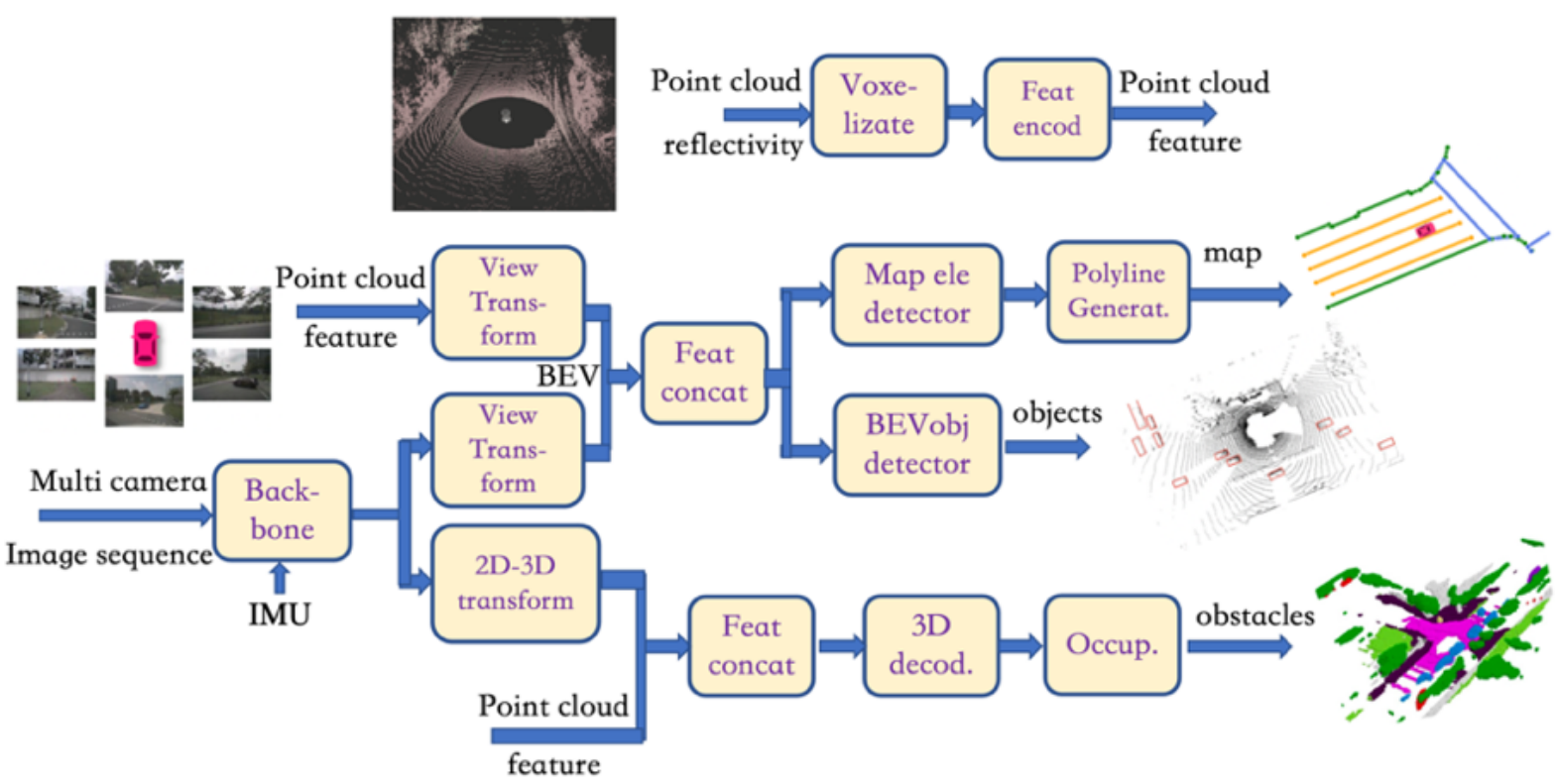

(c) Both camera and LiDAR

Fig. 6 Instances of both BEV and occupancy networks

For camera only input shown in Fig. 6(a), multi camera images are first encoded through the "**backbone**" module, such as EfficientNetor/RegNet plus FPN/Bi-FPN, and then are divided into two paths; on the one hand, image features enter the "**view transform**" module, and BEV feature are constructed through depth distribution or Transformer architecture, and then go to two different heads respectively: one head outputs vectorized representation of map elements through the "**map ele detector**" module (its structure is similar to the transformer based DETR model, also with a deformable attention module and output of the position of key points and the ID of the element they belong to) and the "**polyline generat**" module (it is also a model based on the Transformer architecture with input of those embedded key points, and the polyline distribution model can generate the vertices of the polyline and obtain the geometric representation of map elements), and another head passes through the "**BEV obj Detector**" module to obtain the obj BEV bounding box, which can be implemented using the Transformer architecture or similar PointPillar architecture; on the other hand, in the "**2D-3D transform**" module, the 2-D feature encoding is projected to the 3-D coordinate based on the depth distribution, where the height information is preserved; the camera voxel features obtained then enter the "**3D decod**" module to obtain multi-scale voxel features, and afterwards go into the "**Occupancy**" module for class prediction to generate voxel semantic segmentation.

For LiDAR only input shown in Fig. 6(b), some of modules are the same to Fig. 6(a); firstly, in the "**voxelization**" module, the point cloud is divided into evenly spaced voxel grids, generating a many-to-one mapping between 3D points and voxels; then enter the "**FeatEncod**" module and convert the voxel grid into a point cloud feature map (using PointNet or PointPillar); on the one hand, in the "**view transform**" module, the feature map is projected onto the BEV, where a feature aggregator and a feature encoder are combined, and then BEV decoding is performed in the BEV space, divided into two heads: one head works like Fig. 6(a) going through "**Map Ele Det**" module and "**PolyLine Generat**" module; another head performs BEV object detection through the "**obj det**" module, with a structure similar to the PointPillar model; on the other hand, 3D point cloud feature maps can directly enter the "**3D Decod**" module, obtain multi-scale voxel features through 3D deconvolution, and then perform upsampling and class prediction in the "**Occup.**" module to generate voxel semantic segmentation.

For both camera and LiDAR input shown in Fig. 6(c), most of modules are the same to Fig. 6(a) and 6(b), except the "**Feat concat**" module will concatenate the features from LiDAR path and camera path.

Note: For camera-based occupancy network, it is worth to mention a new paradigm, *Neural Radiance Field* (NeRF) [47] in the fields of computer graphics and computer vision, which is appealing for two unique features: self-supervised and photo-realistic. Rather than directly restoring the whole 3D scene geometry, NeRF generates a volumetric representation called a "radiance field," which is capable of creating color and density for every point within the relevant 3D space.

## 4. Trajectory Prediction

To schedule a safe and efficient navigation, a self-driving car should consider future trajectories of other agents around it. Trajectory prediction is an extremely challenging task which recently gained a lot of attention, which forecasts future state of all the dynamic agents in the scene given their current and past states.

The prediction task can be divided into two parts. The first part is "intention" as a classification task; it can be treated generally as a supervised learning problem, where we need to annotate the possible classification intents of the agent. The second one is "trajectory" which requires to predict a set of possible future locations for an agent in the next $N$ future frames, called as way-points. This builds their interaction with other agents as well as with the road.

Some survey work are given in [10, 12, 34]. Traditionally we classify behavior prediction models based on physics-based, maneuver-based and interactions aware models. *Physics-based* models constitute dynamic equations that model hand-designed motions for different categories of agents. *Maneuver-based* models are the practical models based on the intended motion type of the agents. *Interactions-aware* models are generally an ML based systems that can pair-wise reason every single agent in the scene and produce interaction-aware predictions for all the dynamic agents.

Fig. 7 gives the diagram of the designed prediction model from Cruise.AI [36], which is a L4 autonomous driving startup purchased by GM Motors.

Apparently Fig. 7 illustrates a encoder-decoder framework. In the encoder there is a "**scene encoder**" to process the environment context (map) like Google Waymo's ChauffeurNet (rasterized image as input) or VectorNet (vectorized input) architecture, an "**object history encoder**" to process agent history data (locations), and an *attention-based graph network* to capture joint interactions between agents. To handle the variations of dynamic scenarios, they encode a *mixture of experts* (MoE) into a gating network, for example, there are different behaviors in parking lot, like reverse pull out, pull out & K-turn, parallel parking 2$^{nd}$ attempt, backing & pull out, reverse parallel parking and perpendicular pull out etc.

In the decoder, there a two-stage structure where an *initial trajectory* is generated by a simple regressor, then *refined* by *long term* predictor with "**multi-modal uncertainty**" estimation. To enhance the trajectory predictor, there are some *auxiliary tasks* for training, like "**joint trajectory uncertainty**" estimation and "**interaction detection & uncertainty**" estimation, as well as "**occupancy prediction**".

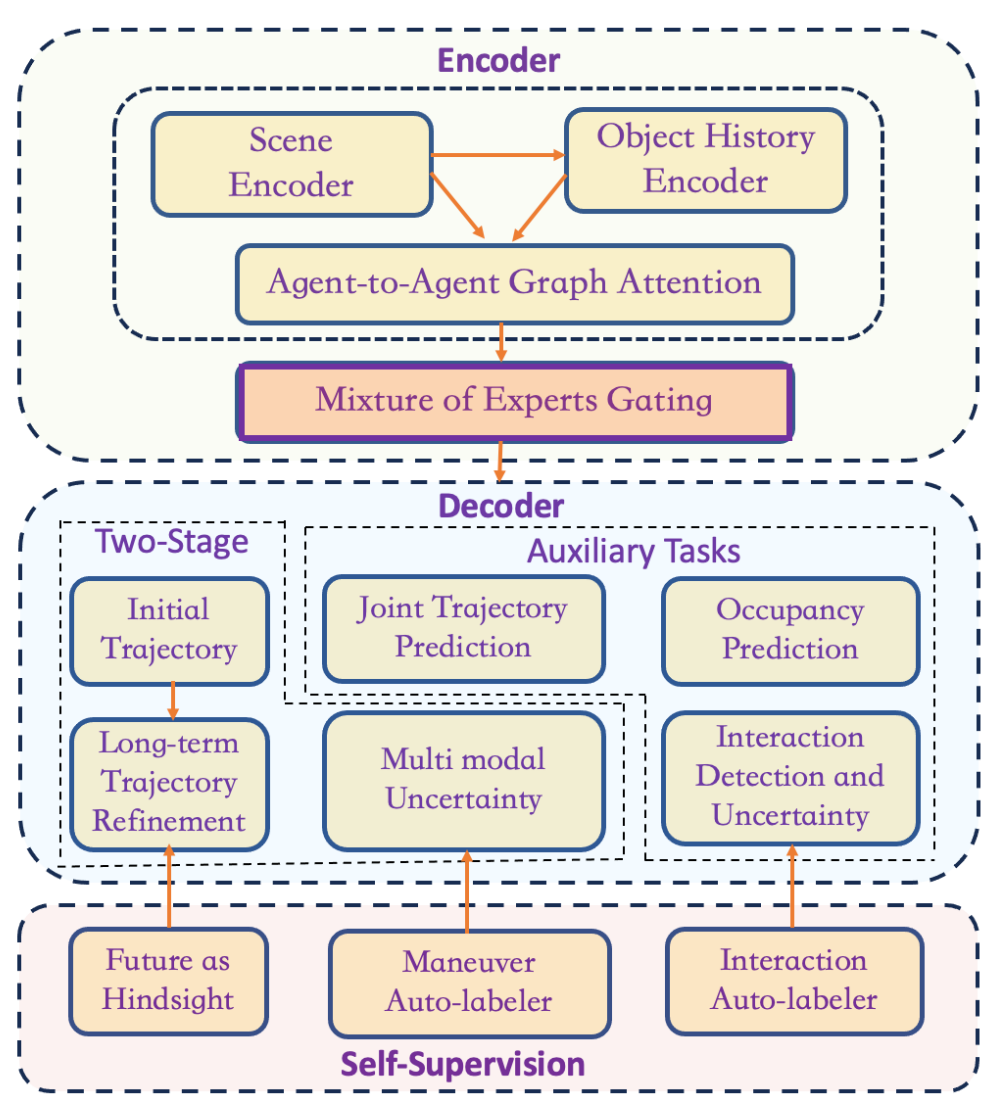

Fig. 7 Prediction model of L4 startup Cruise.AI [36]

A big innovation in this trajectory predictor is its "*self-supervision*" mechanism. Based on the observation "**future as hindsight**", they provide "**maneuver auto-labeler**" and "**interaction auto-labeler**" to generate a large amount of training data for the predictor model.

## 5. Mapping

The map, especially the HD map, is a prior for self-driving. Map building techniques can be classified into *online* and *offline* mapping [24]. In offline mapping, we collect data all in a central location. The data capture comes from a vehicle installed with GNSS, IMU, LiDAR and camera etc. On the other hand, online map building happens onboard using lightweight modules.

All promising mapping technology currently uses the LiDAR as a major sensor especially for the HD map. On the other hand, there are approaches which only use vision sensors to construct the map, like Mobileye's REM or called roadbook, based on visual SLAM and deep learning [35].

HD map generation typically involves collecting high-quality point clouds, aligning multiple point clouds of the same scene, labeling map elements, and updating maps frequently. This pipeline needs a lot of human efforts which constrains its scalability. *BEV perception* [25, 29] offers the potential of online map learning, which dynamically constructs the HD maps based on local sensor observations, appears like a more scalable way to provide semantic and geometry priors to self-driving cars.

Here we introduce a latest work in online mapping, called MachMap [45], which formulates the HD-map construction as the point detection paradigm in the BEV space with an E2E manner. Based on a map compaction scheme, it follows the query-based paradigm integrating with a CNN-based backbone like InternImage, a temporal-based instance decoder and a point-mask coupling head.

The framework of MachMap is illustrated in Fig. 8. It generates 2D features from each of views through image backbone and neck from surrounding images. Then the deformable attention are used to aggregate the 3D feature among different views and average it along z-axis. In the temporal fusion module, the new BEV feature is fused with the one of BEV feature's hidden state.

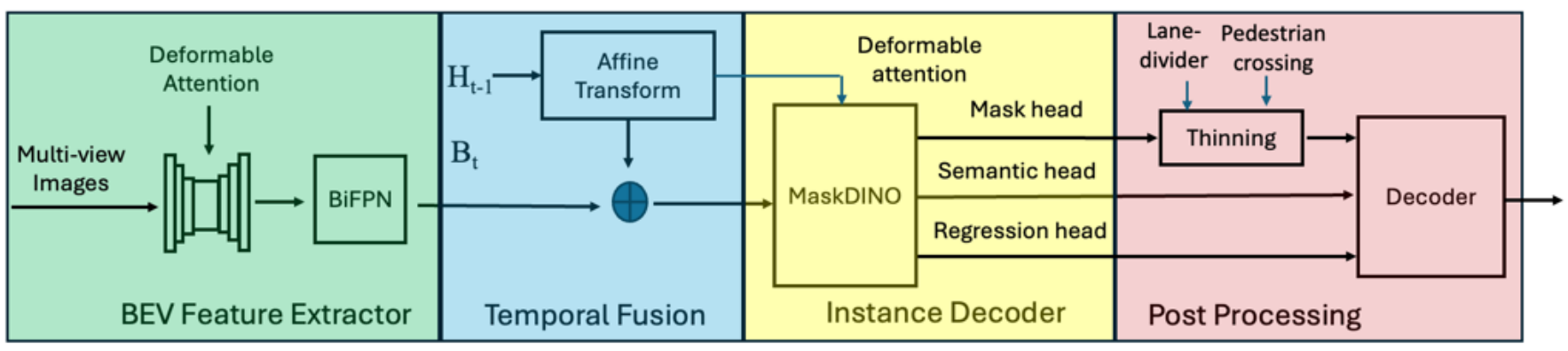

Fig. 8 The framework of MachMap [45]

## 6. Localization

Localizing a self-driving vehicle accurately can make a terrific effect on downstream tasks such as behavioral planning. While it is possible to generate acceptable results with traditional dynamic sensors such as IMUs and GPS, visual based sensors, LiDAR or camera, are apparently suitable for this task, since the localization obtained with such sensors not only rely on the vehicle itself, but also on its surrounding scenes. While both kinds of sensors result into a good localization performance, they also suffer from some limitations [27].

Localization has been on self-driving vehicles through the years, most of time jointly with the mapping aspect, bring about two distinct routes: The first one is SLAM where both the localization and mapping are run simultaneously in a loop; the second splits the localization and mapping while building the map offline.

Recently deep learning has brought data-driven approaches to SLAM, especially the more challenging visual-SLAM, referenced in a survey paper[28]. Here we show an example of transformer-based localization method [48], in which pose queries are updated by interacting with the retrieved relevant information from cross-model features using attention mechanism by a proposed POse Estimator Transformer (POET) module. The localization architecture is illustrated in Fig. 9.

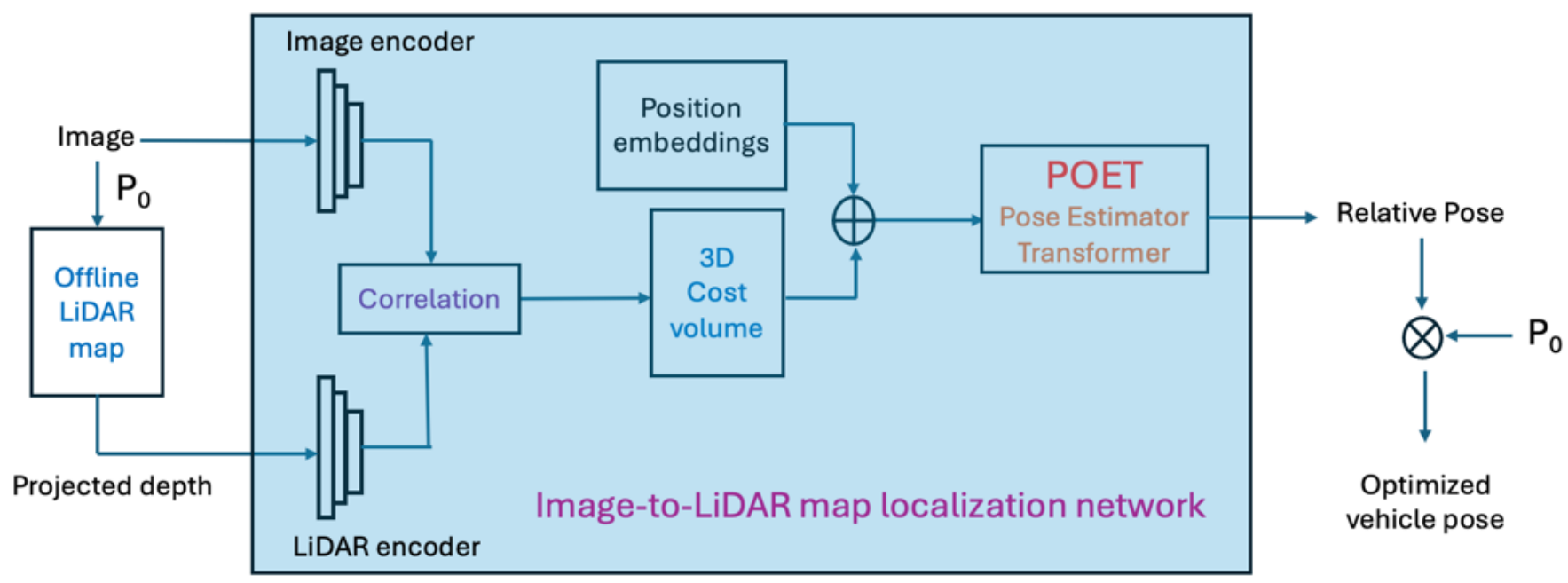

Fig. 9 Map Localization with Transformers [48]

As shown in Fig. 9, the network is fed by a RGB image and a projected depth image, generated by re-projecting the neighboring point clouds in the LiDAR map onto a virtual image plane on a given initial pose. Then, they process with a corresponding encoder to get high-dimensional features respectively. Applying a correlation module, a cost volume between image and LiDAR features is obtained. Then positional embedding is appended to the cost volume and feed the flatted cost volume into the proposed POET module.

## 7. Planning

Most of planning methods, especially behavior planning, are *rule-based*[1, 2, 7-8], which carries the burden to data-driven system exploring and upgrade. The rule-based method planning framework is responsible for calculating a sequence of trajectory points for the low-level controller of the ego-vehicle to track. As a major advantage, the rule-based planning framework is interpretable, the defective module can be recognized when a malfunction or unexpected system behavior happens. The restriction is that it requires many manual heuristic functions.

The *learning-based* planning method has become a tendency in self-driving research[15, 18, 33]. The driving model can learn knowledge via imitation learning and explore driving policy via reinforcement learning. Compared with the rule-based method, the learning-based method deals with vehicle environment interactions more efficiently. Despite its appealing concept, it is difficult or even impossible to find out the reasons when the model misbehaves.

*Imitation learning* (IL) refers to the agent learning policy based on expert trajectory. Each expert trajectory contains a sequence of states and actions, and all "state-action" pairs are extracted to construct datasets. The specific objective of IL is to appraise the most fitness mapping between state and action, so that the agent achieves the expert trajectories as close as possible.

In order to handle the burden for labeled data, some scientists have applied *reinforcement learning* (RL) algorithms for behavior planning or decision making. The agent can obtain some rewards by interacting with the environment. The objective of RL is to optimize cumulative numerical rewards via trial-and-error. By consistently interacting with the environment, the agent gradually acquires knowledge of the optimal policy to arrive at the target endpoint. Training a policy from scratch in RL is frequently time consuming and difficult. Combining RL with other methods such as IL and curriculum learning may serve as a viable solution.

In recent years deep learning (DL) techniques has provided powerful solutions to the behavior planning problem through the wonderful properties of deep neural networks (DNNs): function approximation and representation learning. DL techniques enable the scaling of RL/IL to previously intractable problems (e.g., high-dimensional state spaces).

Here a two-stage occupancy prediction-guided neural planner (OPGP) [46] is presented, which integrates joint predictions for future occupancy and motion planning with prediction guidance, illustrated in Fig. 10.

In the first stage of OPGP, an integrated network is established upon Transformer backbones. The visual features are a combination from the historical occupancy grids and rasterized BEV roadmap representing the spatial-temporal status of traffic participants under a specific scenario. The vectorized context initially concerns the dynamic context from actors centered on the autonomous vehicle. Occupancy predictions for all types of traffic participants are output simultaneously taking into account interaction awareness for both visual features and vectorized context. Meanwhile, encoded scene features and occupancy are shared and conditionally queried in the planner head, which conducts multi-modal motion planning.

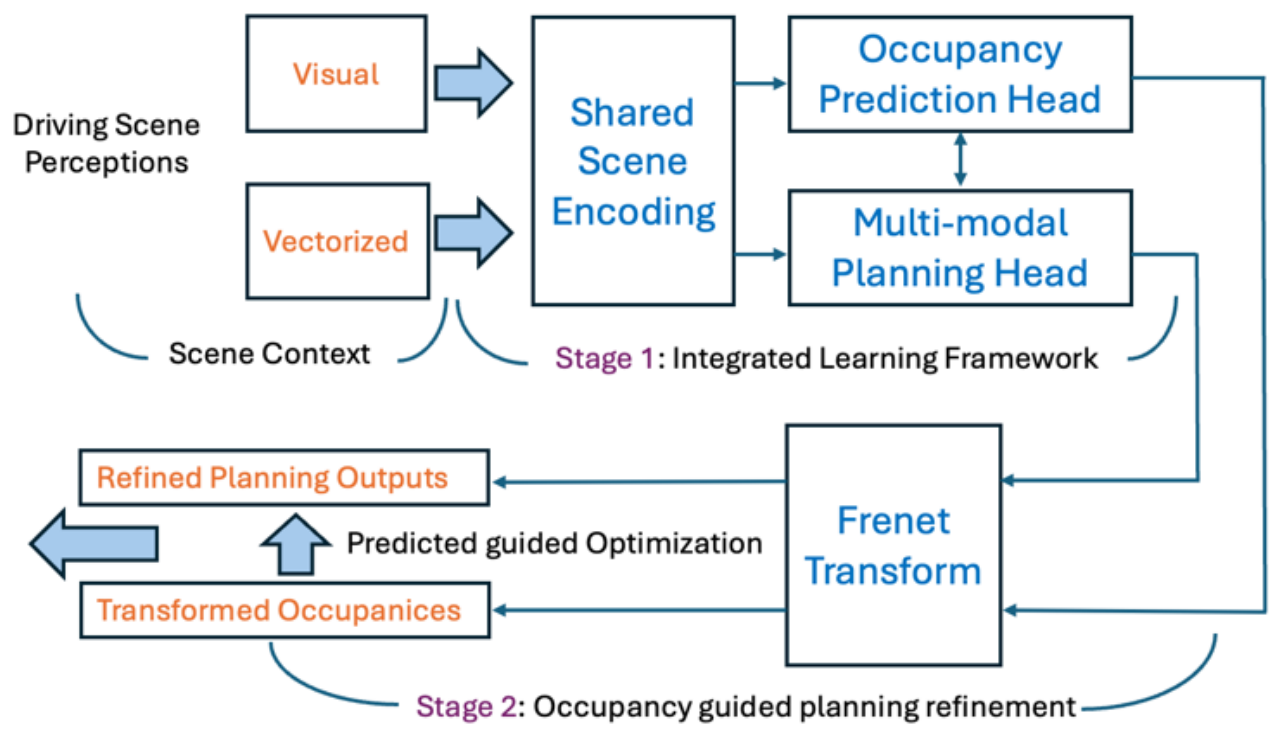

Fig. 10 A two stage OPGP [46]

The second stage of OPGP focuses on modeling explicit guidance from occupancy prediction for refinement in an optimization-feasible manner. More specifically, they construct an optimization pipeline in Frenet space （it is a moving right-handed coordinate system determined by the tangent line and curvature）for planning refinement using transformed occupancy predictions.

## 8. Control

Vehicle control is relatively mature and the classical control theory plays the main role, compared with other modules in the autonomous driving pipeline, such as perception and planning [20, 21]. However, deep learning methods have made great promise in not only obtaining excellent performance for various non-linear control problems, but also in extrapolating previously learned rules to new scenarios. Therefore, the application of deep learning for self-driving control is becoming more and more popular [13].

There are a diversity of sensor configurations; whilst some people aim to control the vehicle with visual vision only, others make use of lower dimensional data from ranging sensors (LiDAR or radar), and some utilize multi-sensors. There are also variations in terms of the control objective, some formulate the system as a high-level controller to provide the target, which is then achieved through a low-level controller, often using classical control techniques. Others aim to learn driving end-to-end, mapping observations directly to low-level vehicle control interface commands.

The vehicle control can be broadly divided into two tasks: lateral and longitudinal control. Lateral control systems aim to control the vehicle's position on the lane, and to realize other lateral actions such as lane changes or collision avoidance maneuvers. In the deep learning domain, this is typically achieved by capturing the environment using the images/point clouds from on-board cameras/LiDARs as the input to the neural network.

Longitudinal control manages the acceleration of the vehicle such that it keeps the desirable velocity on the lane, maintains a safe distance from the preceding vehicle, and avoids rear-end collisions. While lateral control is typically achieved through visual input (cameras), the longitudinal control relies on measurements of relative velocity and distance to the preceding/following vehicles. This means that ranging sensors such as RADAR or LIDAR are more commonly used in longitudinal control systems.

In this session, we describe an end-to-end (E2E) driving models with semantic visual map and cameras [16]. Human-like driving is realized using adversarial learning where a generator imitates the human driver and a discriminator to make it human-like.

The training data (named as "Drive360 dataset") is captured by a front-facing camera and a rendered TomTom route planning module. Then the dataset is augmented offline with HERE map data to provide a synchronized semantic map information.

For the basic E2E driving model, a sequence of past images and map renders are recorded, and the action is predicted. The architecture of the network is shown in Fig. 11(a): The image is fed through a visual encoder, and the output latent variable $z_I$ is further fed into an LSTM, which results in a hidden state $h$; the map renders are also processed in a visual encoder, generating another latent variable $z_M$; these three variables are then concatenated to predict the actions.

The naïve approach with additional the semantic map information is called the late fusion approach, which diagram is given in Fig. 11(b): A vector embeds all the semantic map information, processed by a fully connected network, which output latent variable $z_n$ is concatenated with $z_I$, $z_M$ and $h$.

A new method is proposed that promotes output class probabilities of a segmentation network based on semantic map information, which complete architecture is shown in Figure 11(c). This approach uses a semantic segmentation network that obtains a confidence mask for all 19 classes, which is then pushed using a soft attention network to generate a 19 class attention vector.

When training the driving model, the decision making problem is regarded as a supervised regression problem to matching action sequences (called *drivelets*). A *Generative Adversarial Network* (GAN) is used to formulate the imitation learning problem, where the generator is the driving model, and the discriminator recognizes if a drivelet is human-like.

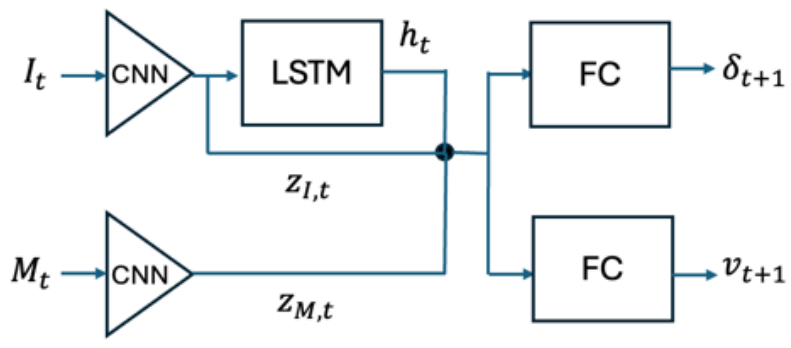

(a) Basic E2E driving model

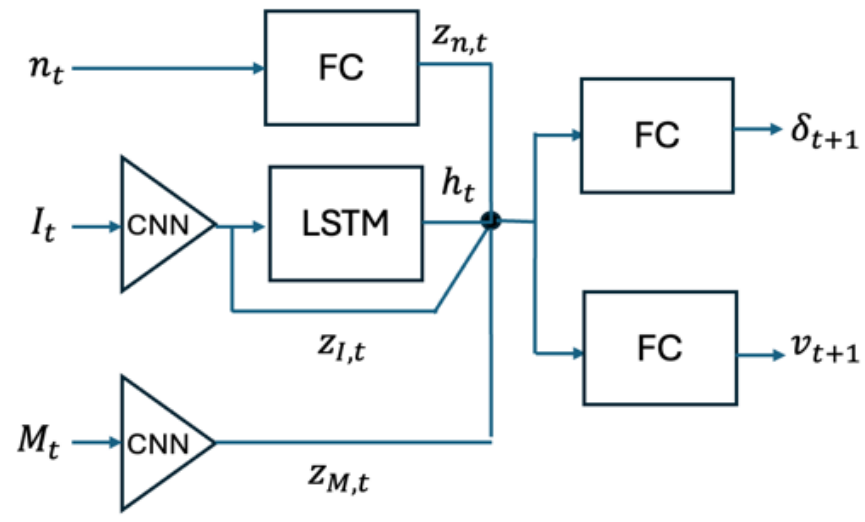

(b) E2E driving model with semantic map data

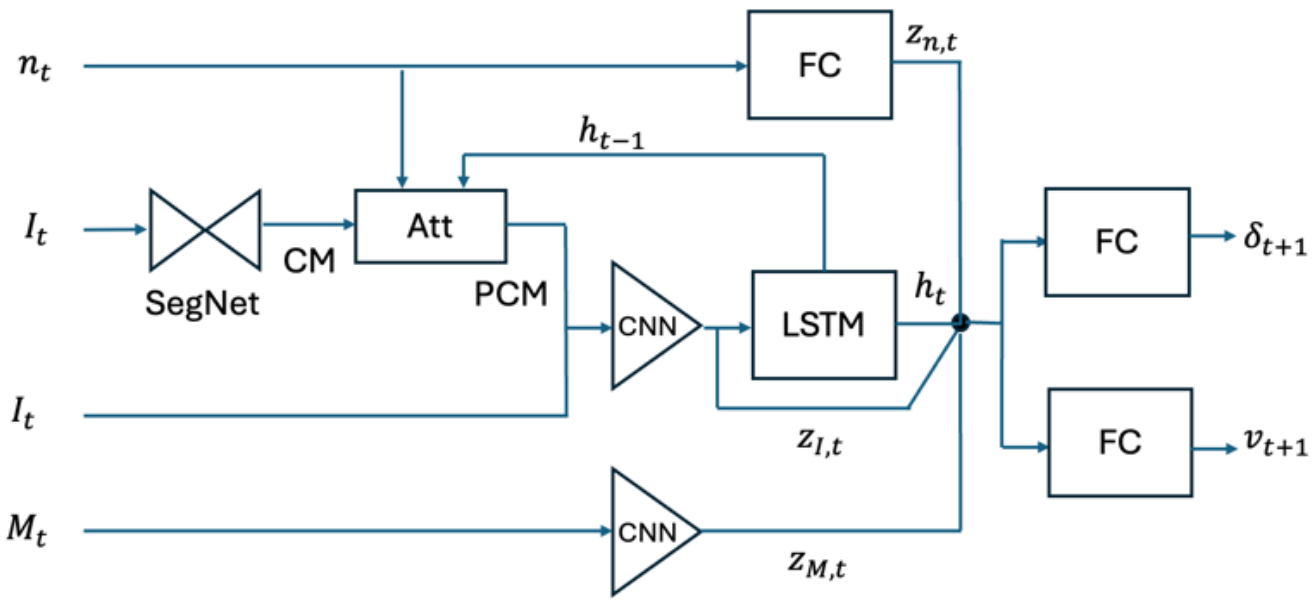

(c) E2E driving model with semantic map data and attention

Fig. 11 Three E2E driving model frameworks [16]

## 9. V2X

Benefiting from the better building of communication infrastructure and developing communication technology such as *Vehicle-to-Everything* (V2X) communication, vehicles could exchange their messages in reliable manners, which enables the collaboration among them[4,11]. Cooperative driving leverages *Vehicle to Vehicle* (V2V) and *Vehicle to Infrastructure* (V2I) communication technologies aiming to carry out cooperative functionalities: (i) cooperative sensing and (ii) cooperative maneuvering.

There are some generic cooperative driving scenarios: smart parking, lane change and merge as well as cooperative intersection management. *Vehicle platooning*, also known as convoy driving, is the practice of driving a group of two or more consecutive vehicles nose-to-tail on the same lane with small inter-vehicle spacings typically less than 1 second at the same speed, which is an main use-case leading towards cooperative autonomous driving [26].

Valuable research efforts using either centralized or decentralized approaches have focused on coordinating CAVs in intersections and merging at highway on-ramps. In *centralized* approaches, there is at least one task in the system that is globally decided for all vehicles by a single central controller. In *decentralized* control, each vehicle determines its own control policy based on the information received from the other vehicles on the road, or some coordinator.

The decentralized method could be categorized as three types: negotiation, agreement and emergent. Most representative *negotiation* protocols are: Contract Net for collaborative problems, and auctions for competitive ones. *Agreement* during the coordination process will outcome both the set of admissible moves and even a dynamic re-determination of the goals. *Emergent* rather makes every vehicle behaves in a selfish way according to its goals and the perceptions, for example, game theoretic or self-organizing.

Instead of individual perception, collaborative or cooperative perception, which leverages the interaction between multiple agents to improve perception in autonomous driving, has received considerable critical attention [31]. As deep learning methods have been widely applied to self-driving perception, efforts to improve the ability and reliability of collaborative perception systems are steadily increasing.

According to the message delivered and the collaboration stage, the collaborative perception scheme can be broadly separated into early, intermediate, and late collaboration. *Early collaboration* employs the raw data fusion at the input of the network, which is also known as *data-level fusion* or *low-level fusion*. Considering the high bandwidth of early collaboration, some works propose *intermediate collaborative* methods to balance the performance-bandwidth trade-off. *Late collaboration* or *object-level collaboration* employs the prediction fusion at the network. The challenging issues in collaborative perception are: calibration, localization of vehicles, synchronization and spatial registration etc.

Here we propose a hierarchical V2X sensing platform shown in Fig. 12. Time sync information convey the time difference between data from different agents. To be flexible, the data container is preferred to keep a temporal window, for example, 1 second (10 frames for LiDAR/radar and 30 frames for camera). Pose information is required for spatial registration, acquired from vehicle localization and calibration, mostly that is based on online mapping or matching with information from HD map built offline.

We assume the sensors are cameras and LiDARs. The neural network model can process the raw data to output intermediate representation (IR), scene segmentation and object detection. To unify the collaboration space, the raw data are mapped to BEV (bird eye view) and processed results are also located in the same space.

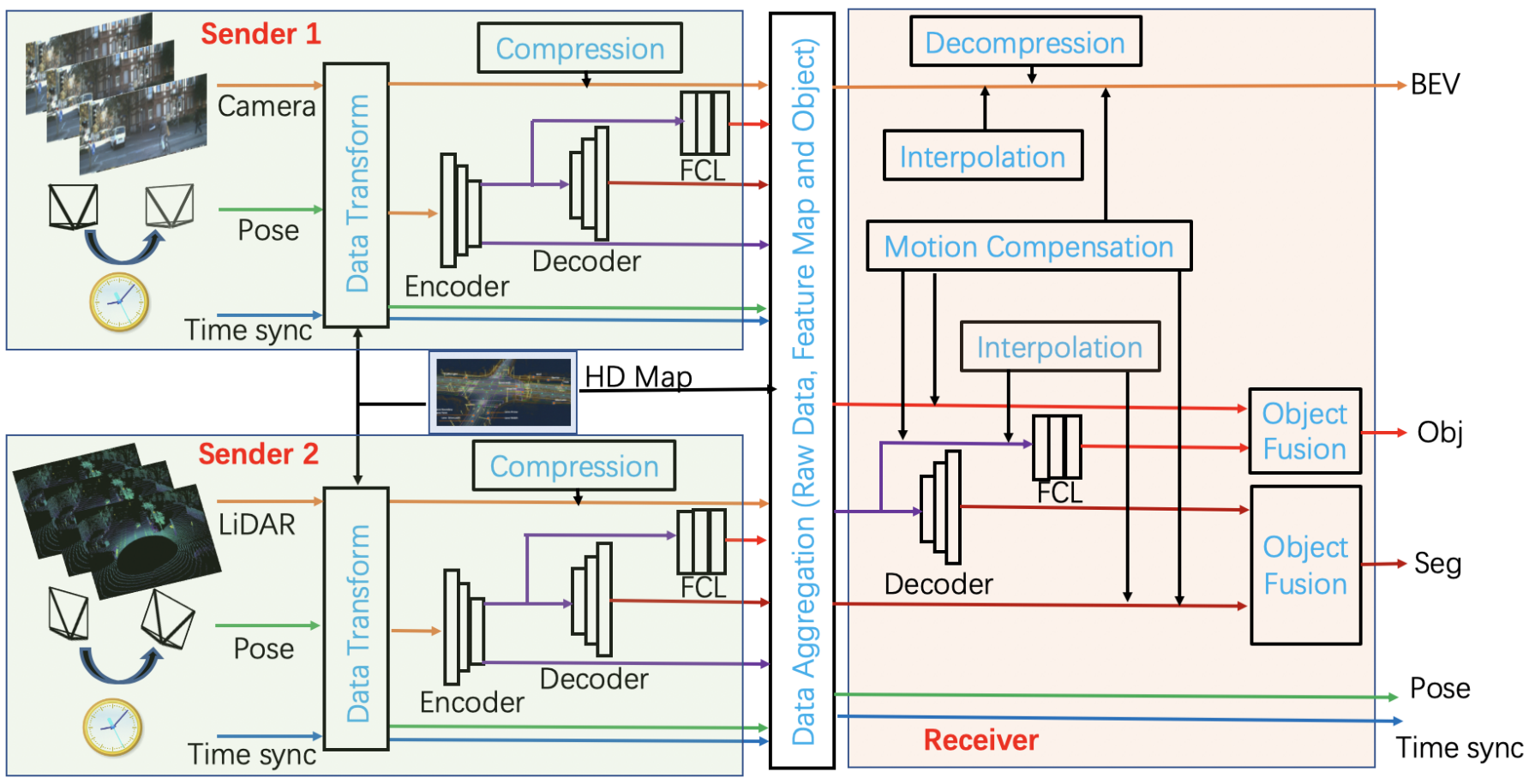

Fig. 12 Diagram of collaborative sensing in V2X

To keep a limited scale space, multiple layers in IR are reserved, such as 3, which allows flexible fusion of different data resolution. Collaborative sensing of V2X needs more work in the receiver to integrate the information from other vehicles and roadsides, fusing IR, segmentation and detection channels respectively. The fusion module could be CNN-based, Transformer-based or graph neural network (GNN)-based.

Note: FCL stands for fully connected layer, modules "**compression**" and "**decompression**" are required for raw data, modules "**interpolation**" and "**motion compensation**" are useful at the receiver based on the time sync signal and relative pose based on online mapping/localization/HD map (which is built offline).

## 10. Simulation

The physical testing on closed or public roads is unsafe, costly, and not always reproducible. This is where testing in *simulation* helps fill the gap, however the problem with simulation testing is that it is only as good as the simulator used for testing and how representative the simulated scenarios are of the real environment[17].

An ideal simulator is the one that is as close to reality as possible. However, this means it must be highly detailed in terms of 3D virtual environment and very precise with lower level vehicle calculations such as the physics of the car. So, there is a trade-off between the realism of the 3D scene and the simplification of the vehicular dynamics.

In general, the learned driving knowledge in virtual scenes is transferred to the real world, so how to adapt driving knowledge learned in simulation to reality becomes a critical issue. The gap between the virtual and real worlds is commonly referred as "*reality gap*". To handle this gap, various methods are put forward classified as two categories: knowledge transfer from simulation to reality (*sim2real*) and learning in digital twins (*real2sim*) [44].

There are gradually 6 kinds of methods developed in sim2real, including curriculum learning, meta-learning, knowledge distillation, robust reinforcement learning, domain randomization, and transfer learning. Digital twin-based methods aim to construct a mapping of real-world physical entities in a simulation environment using the data from sensors and physical models to achieve the role of reflecting the entire lifecycle process of corresponding physical entities, like AR (augmented reality) and MR (mixed reality).

Even though testing self-driving systems in simulation is relatively cheap and safe, the safety critical scenarios generated for evaluation should be more important to manage the risk and reduce the cost[22]. Actually the safety critical scenarios are rare in the real world, so various methods to generate those scenario data in simulation are put on research, divided into three types: *data driven generation* that only leverages information from the collected datasets to generate scenarios, *adversarial generation* that uses the feedback from the autonomous vehicle that is deployed in the simulation, and *knowledge-based generation* that leverages the information mainly from external knowledge as constraints or guidance to the generation.

Here we report a latest neural sensor simulation platform [49] - UniSim, built by Waabi, U. of Toronto and MIT. UniSim takes a single recorded log captured by a sensor-equipped vehicle and converts it into a realistic closed-loop multi-sensor simulation, as a editable and controllable digital twin. A feature field refers to a continuous function f that maps a 3D point and a view direction to an implicit geometry and a feature descriptor, which is defined as neural feature field (NFF). NFFs naturally support composition, enabling the combination of multiple relatively simple NFFs to form a complex field.

UniSim is a neural rendering [47] closed-loop simulator that jointly learns shape and appearance representations for both the static scene and dynamic actors from the sensor data captured from a single pass of the environment. To better handle extrapolated views, learnable priors are incorporated for dynamic objects, and a convolutional network is leveraged to complete unseen regions.

Shown in Fig. 13, the 3D scene in UniSim is divided into a static background (grey) and a set of dynamic actors (red). The neural feature fields are queried separately for static background and dynamic actor models, and volume rendering is performed to generate neural feature descriptors. The static scene is modeled with a sparse feature-grid and use a Hypernet to generate the representation of each actor from a learnable latent. Finally a CNN is used to decode feature patches into an image.

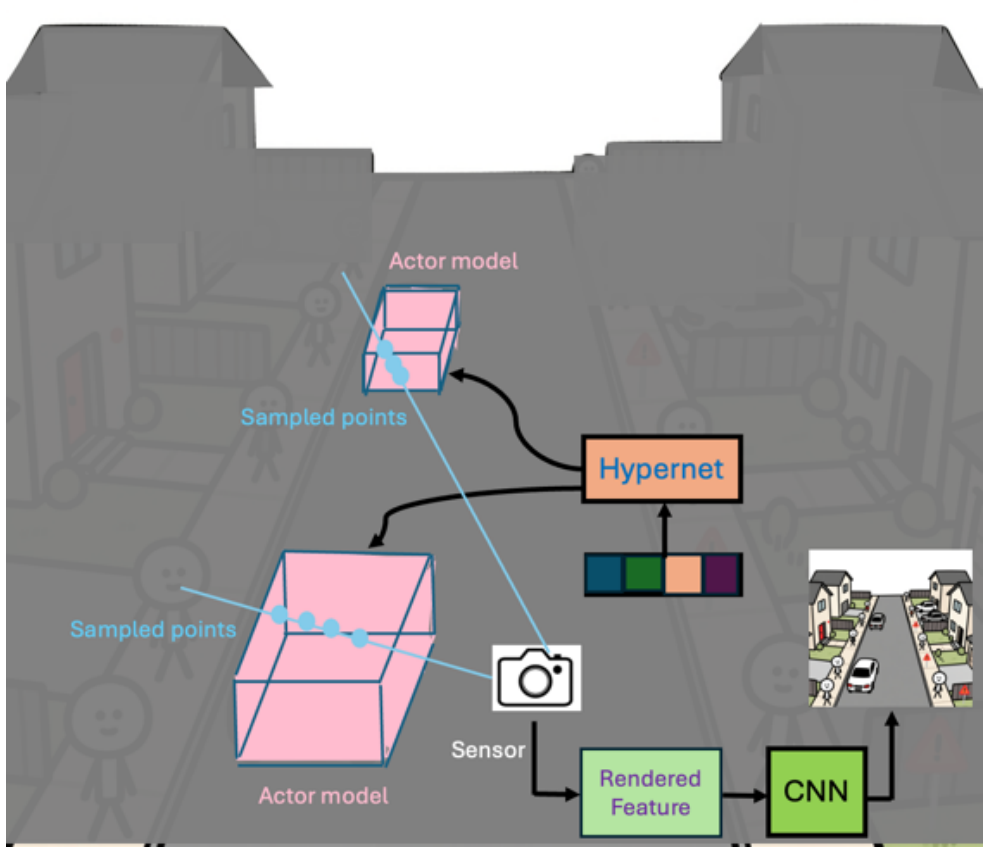

Fig. 13 Overview of the sensor simulator UniSim [49]

Note: An emerging class of generative models, called *diffusion models* [50], with a generic pipeline involving a forward process and a backward process to learn the data distribution as well as a sampling procedure to generate novel data, have gained significant attention in computer vision. Recently it has become increasingly popular in image-to-image, text-to-image, 3D shape generation, human motion synthesis, video synthesis and so on. It is expected diffusion models synthesize imaginable driving scenario contents for simulators in autonomous driving.

## 11. Safety

Safety is the major concern in real-world deployment of autonomous driving systems (ADS) [19, 23]. Except for the classical attacks to sensors and cyber systems, AI or machine learning (including deep learning)-based systems especially need take into account new safety issues brought about by the neural network's innate vulnerability to adversarial attacks from adversarial examples.

ISO 26262 Road Vehicles—*Functional Safety* is the widely used safety guideline standard, applicable only for alleviating known unreasonable risks related to known failure of components (i.e. known unsafe scenario), but not coming up to AV driving risks because of complex environment variants and how the ADS responses to them, while there is no technical failing in the vehicle.

Currently adversarial defenses can be categorized into proactive and reactive methods. The proactive method concentrates on ameliorating the robustness of the targeted AI models, while the reactive ones targets at detecting counter adversarial examples before they are fed into models. There are five main types of *proactive defense* methods: adversarial training, network distillation, network regularization, model ensemble, and certified defense. The *reactive defenses* try adversarial detection and adversarial transformation.

Interpretability is an arising issue from the black-box nature of deep neural networks. Straightforwardly, it should give a human-perceivable interpretation for the behavior of a deep learning model. An interpretation procedure can be separated into two steps: an extraction step and an exhibition step. The extraction step obtains an intermediate representation, and the exhibition step presents it for humans to digest in an easy way. In autonomous driving, visualizing the feature map in the model backbone or managing the loss of the decoder head output, are viable means to enhance the interpretability.

To supply safety guarantee, substantial amount of verification and validation (V&V) is expected against the scale of scenarios in real world that ADS will face. A regular policy of V&V to maximize scenario coverage, is to affirm ADS in simulated huge amount of generated scenario samples. The methods to ensure reasonable coverage are classified as two groups: scenario sampling-based and formal methods.

The *scenario sampling* method is the major one for AI safety control, including *testing-based sampling* to maximize scenario coverage at minimum effort and *falsification-based sampling* to find corner cases that are paid more attention by developers, such as safety-critical scenarios.

ISO 21448 *Safety of the Intended Functions* (SOTIF) proposes a qualitative goal, describing in high-level how to minimize known and unknown unsafe scenario consequences for ADS function design[23], shown in Fig. 14. The sample based methods are less biased and more exploratory in finding unknown unsafe scenarios, and the move from unknown to known behaves in a way that all sampled scenarios are within a consistent simulation environment and the same fidelity level.

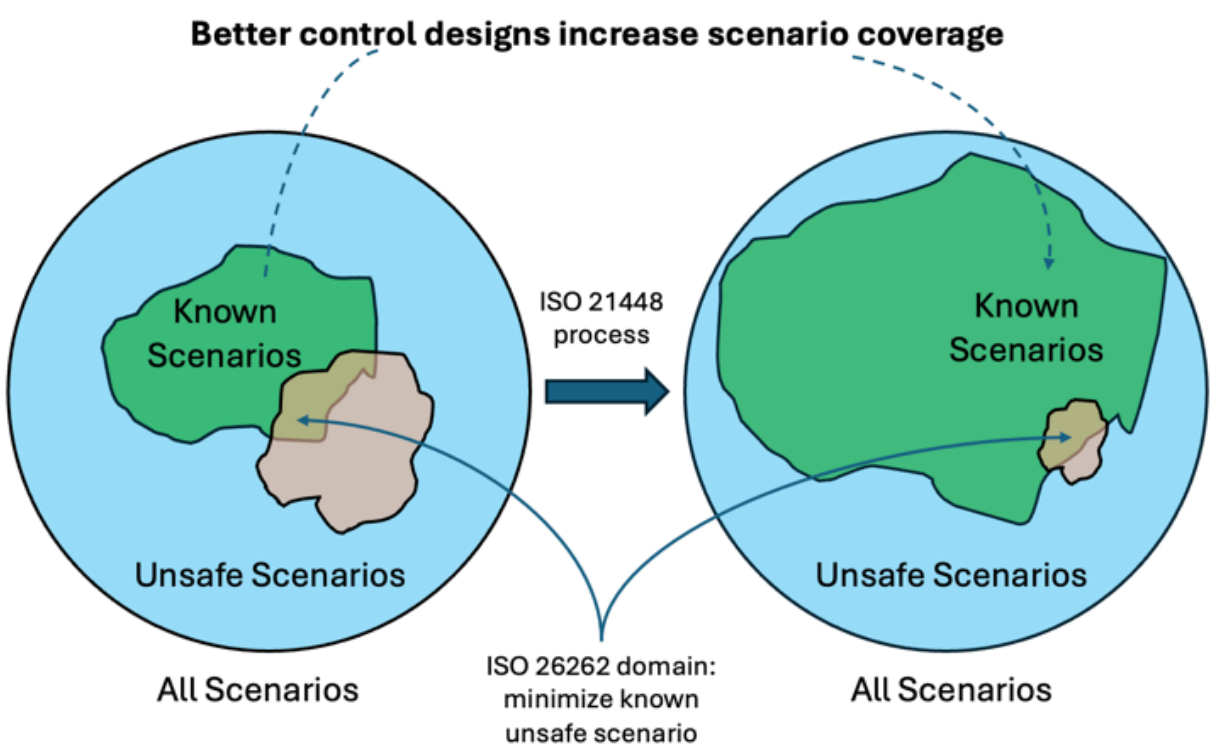

Fig. 14 The goal of SOTIF[23]

Popularly used *formal methods* in AV safety consists of model checking, reachability analysis and theorem proving. *Model checking* comes from software development to guarantee that the software behavior adheres to the design specifications. When safety specification is described in axioms and lemmas, then *theorem proving* is performed to testify safety using worst case assumptions. *Reachability analysis*, owing to producing safety statements for dynamical systems, estimates the characteristic of dynamic driving task (DDT), such as Mobileye's safety model RSS (*Responsibility-Sensitive Safety*) and Nvidia's safety model SFF (*Safety Force Field*).

## 12. Data Closed Loop

Capturing data from the vehicle, selecting valuable data, annotating them, training/finetuning the expected model, validating and deploying the target model to the vehicle etc., constitute a *data closed loop* for research & development (R&D) of autonomous driving[37-41], shown in Fig. 15.

As an automated driving development platform, the data closed loop should include both the client vehicle end and server cloud, implementing data collection and preliminary screening at the vehicle side, mining based on active learning in the cloud side database, automatic tagging, model training and simulation testing (simulation data might also join model training), and model deployment back to the vehicle side. Data selection/screening and data labeling/annotation are key modules that determine the efficiency of the data closed-loop.

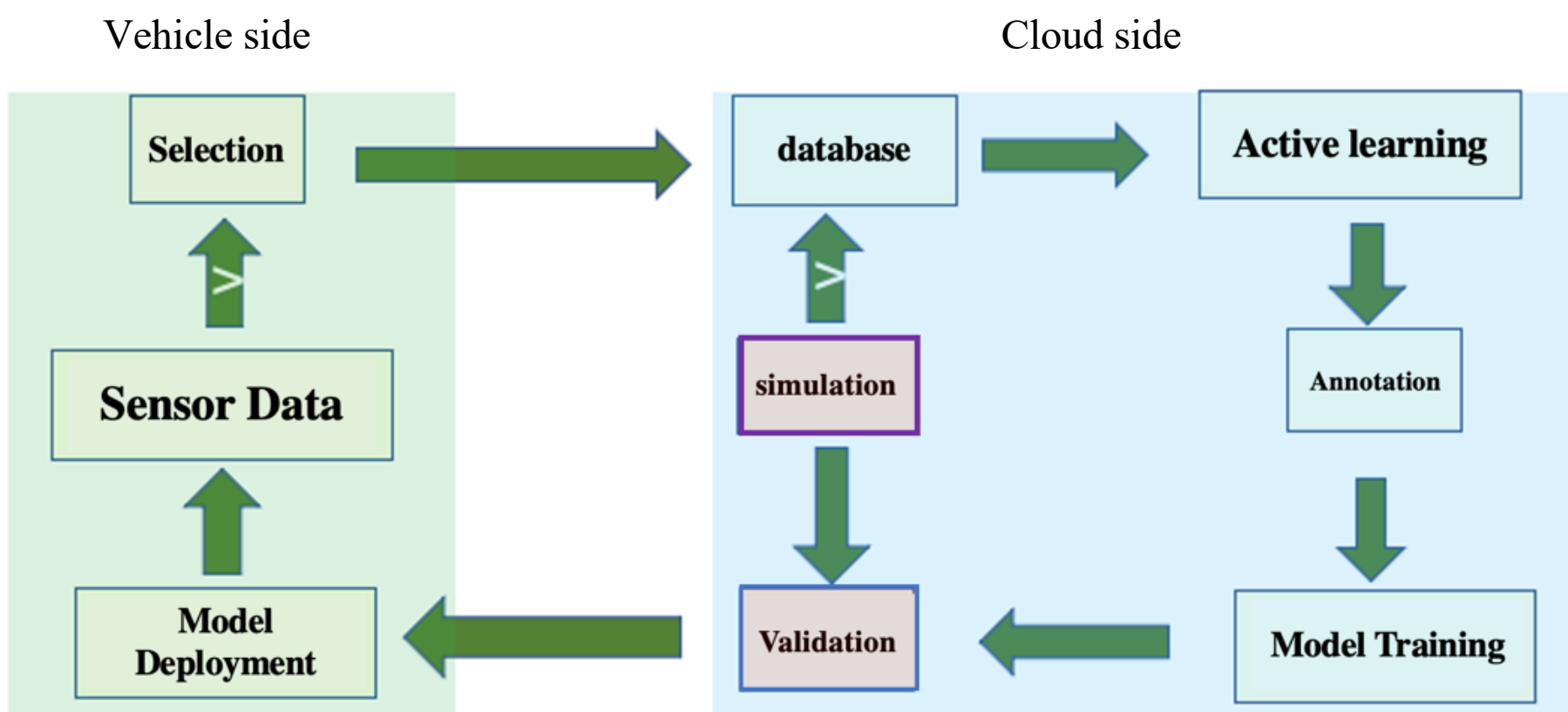

Fig. 15 The data closed loop in autonomous driving R&D

### 12.1. *Data Selection*

Tesla was the first company to explicitly propose the *data selection* strategy on mass-produced vehicles, known as the online "*shadow mode*". It can be seen that data selection is divided into two ways: one is online style, where the trigger mode for data collection is set on the human driven vehicle, which can collect the required data most economically; this style is mostly used during the mass production and business stages (note: business vehicles with safety operators usually manually trigger the collection directly); another is the database model, which generally adopts a data mining model to screen incremental data in cloud server; this kind of model is commonly used in the R&D stage, and even the data collected in the mass production stage will be subjected to secondary screening in the server-side data center; in addition, in cases where there is a significant lack of data for known scenarios or targets, a "content search" mode can also be set on the vehicle or server side to search for similar object, scene or scenario data to enhance the diversity of training data and the generalization of the model.

In autonomous driving, there are equivalent or similar concepts for *corner cases*, such as anomaly data, novelty, outliers, and *out of distribution* (OOD) data. The detection for corner cases can be divided as online or offline mode. The *online* mode is usually used as a safety monitoring and warning system, while the *offline* mode is generally used for developing new algorithms in the laboratory to choose suitable training and testing data. Corner cases can be defined at several different levels: 1) pixel/voxel; 2) domain; 3) object; 4) scene; 5) scenario. The corner case of the last scenario level is often not only related to perception, but also involves prediction and decision planning.

Here we propose a data selection framework both online and offline shown in Fig. 16. For online mode given in Fig. 16(a), we adopt multiple screening paths, such as content search, shadow mode, driving operations, and one-class classification. In content search mode, based on a given query, the "**Scene/Scenario Search**" module extracts features (spatial or temporal information) from images or consecutive frames for pattern matching to discover certain objects, contexts, or traffic behaviors, such as motor cycles appearing on the streets at night, large trucks on highways in adverse weather, vehicles and pedestrians in roundabouts, lane changing at high way, and U-turn behavior at street intersections etc.

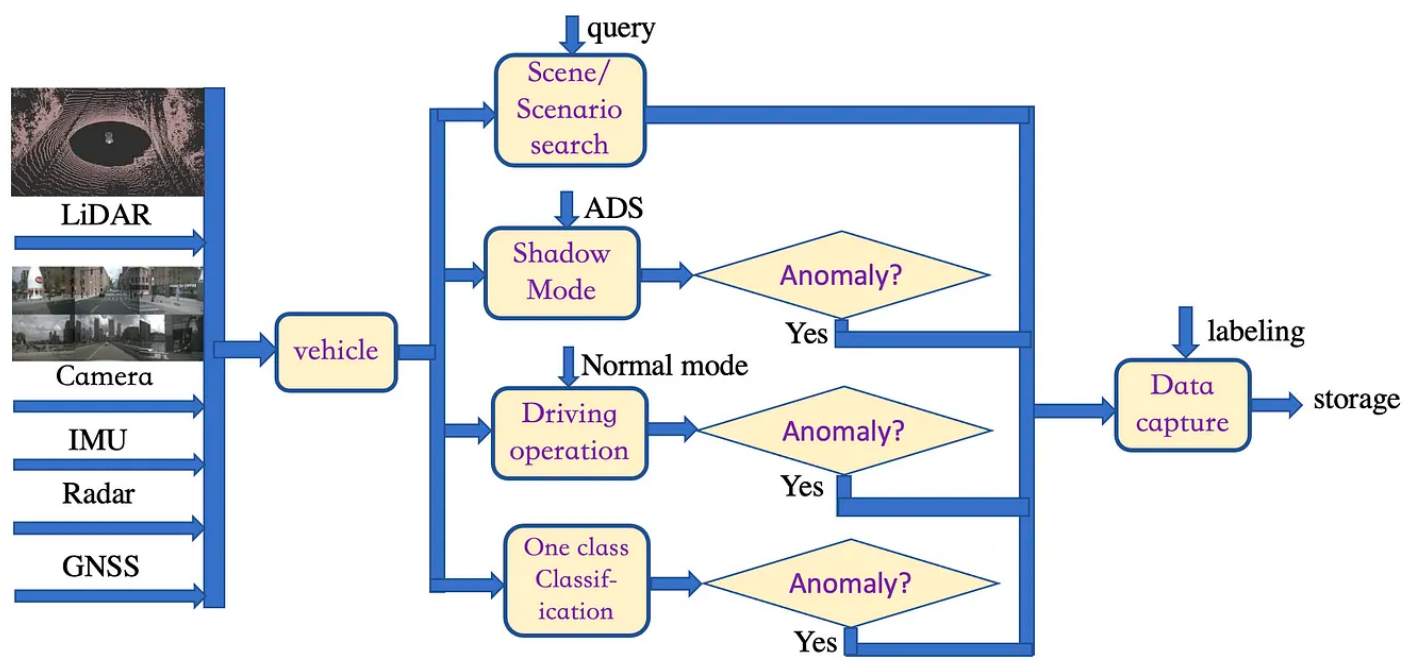

(a) Online mode on the client vehicle

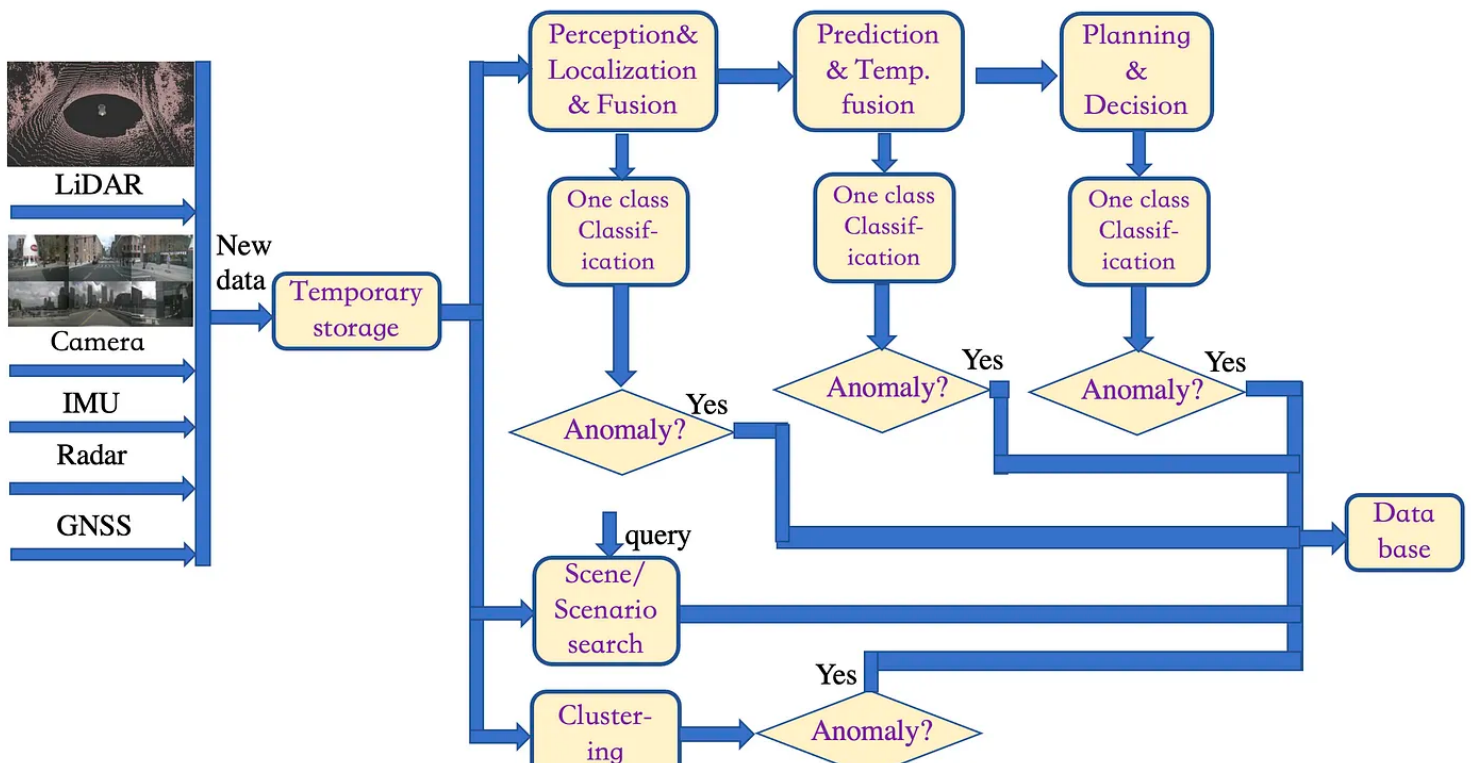

(b) Offline mode on the server cloud

Fig. 16 Data selection both online and offline

The "**Shadow Mode**" module makes judgments based on the results of on-board autonomous driving system (ADS), such as detecting object matching errors in different cameras in the perception module, shaking or sudden disappearing of continuous frame detection, and strong lighting changes at entrance and exit of the tunnel, as well as the behavior of vehicles cutting in but accelerating or vehicles cutting out but decelerating in decision planning, anomaly cases such as detecting obstacles at the front but not trying to avoid them, and approaching and almost colliding with vehicles detected by rear side cameras during lane changing etc.

The "**Driving Operation**" module will detect anomaly from data such as yaw rate, speed, etc. obtained from the vehicle's CAN bus, such as weird zig-zag phenomenon, excessive acceleration or braking, large angle steering or turning angle, even triggering Abrupt Emergent Breaking(AEB).

The "**one-class classification**" module generally trains anomaly detectors for the data in perception, prediction, and planning, and is kind of a generalized data-driven "shadow mode"; It trains one-class classifiers based on the normal driving data, i.e. perception features, predicted trajectories and planned paths, respectively; For lightweight running at the vehicle end, the One-Class SVM model is used. Finally, label each captured data in "**Data Capture**" module according to its collection path.

For offline mode given in Fig. 16(b), we also choose multiple paths for data screening. No matter whether the new data is collected from a R&D data collection vehicle or a mass-production sold user vehicle, it will be stored in a "**temporary storage**" hard disk for the second selection. Similarly, another "**scene/scenario search**" module directly retrieves data according to query which defines a certain kind of scenario. The algorithm/model applied will be larger in size and more time-consuming in computation, without the real time limitation. In addition, data mining technology can be used. The "**clustering**" module will perform some unsupervised grouping methods or density estimation methods to generate scenario clusters. So certain data far away from the cluster centroids creates anomaly.

To further screen data, an autonomous driving software can run step-by-step on it (like the LogSim style) and anomaly can be detected based on a serie of designed checking points. Here, autonomous driving adopts a modular pipeline, which includes the "**perception/localization/fusion**" module, the "**prediction/time-domain fusion**" module, and the "**planning & decision-making**" module. Each module's output is a checking point for anomaly detection through a "**one-class classification**" module, which model architecture is different from that at the vehicle side. This kind of anomaly detector is more complicated because there is no real-time restriction. On the server side, a deep neural network for one-class classification can be performed. This is a kind of "shadow mode" offline.

The architecture of "**perception/localization/fusion**" module is similar to that given in Fig. 6. The "**prediction/temporal fusion**" module serves as an additional output head, which diagram is shown in Fig. 17. The features enter the "**Temporal Encod**" module, which architecture can be designed either similar to the RNN (GRU or LSTM) model or the graph neural network (GNN)-based interaction modeler, fusing multiple frame features. The "**Motion Decod**" module comprehends spatiotemporal features similar to the BEVerse model, and outputs predicted trajectories.

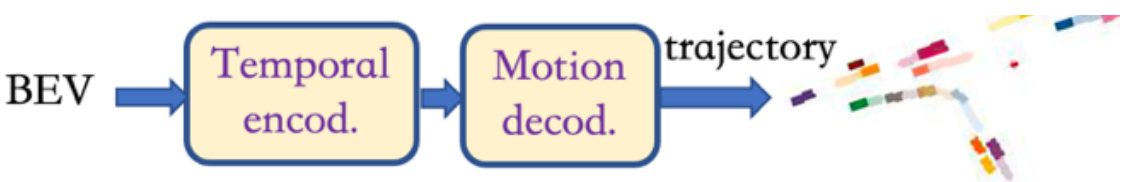

Fig. 17 Prediction based on BEV perception features

On the basis of perception and prediction, we design the block diagram of the planning and decision-making algorithm similar to ST-P3, shown in Fig. 18. We choose a sampling based planning approach, based on the spatio-temporal BEV features output from prediction, train a cost function in the "**Plan Decod**" module to calculate various trajectories generated by the sampler, and find the one with minimal cost in the "**ArgMin**" module. The cost function includes some terms of safety (avoiding obstacles), traffic rules, and trajectory smoothness (acceleration and curvature). Finally, the global loss function is optimized for the whole perception-prediction-planning pipeline.

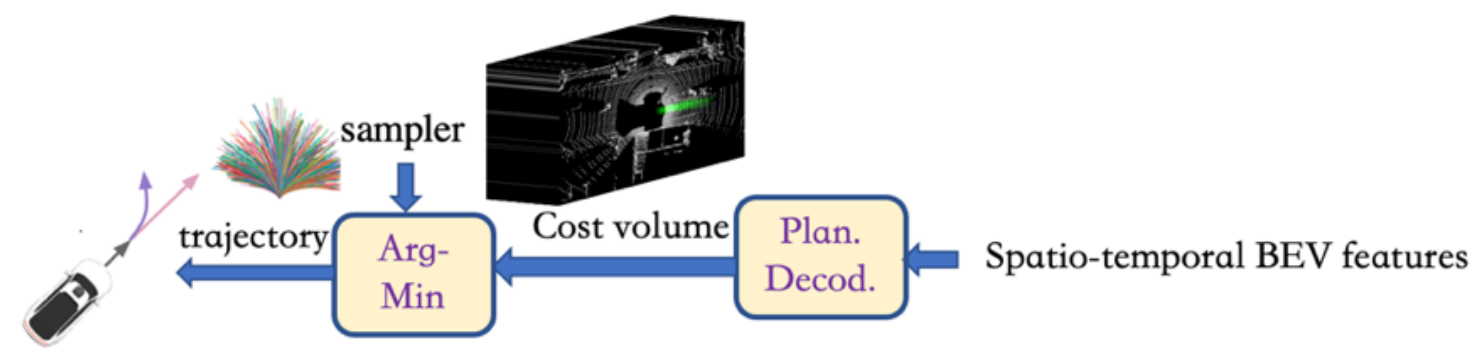

Fig. 18 Planning and decision making based on BEV perception plus prediction

### 12.2. *Data Annotation*

The task of *data annotation* is classified as R&D stage and mass production stage: 1) The R&D stage mainly involves the data collection vehicle of the development team including LiDAR, so that LiDAR can provide 3D point cloud data to the camera's image data to provide 3D ground truth values. For example, BEV (birds' eye view) visual perception requires obtaining BEV output from 2D images, which involves perspective projection and speculation of 3D information; 2) In the mass production stage, data is mostly provided by customers of passenger vehicles or operational customers of commercial vehicles. Most of them do not have LiDAR data or are only 3D point clouds from a limited FOV (such as forward facing). Therefore, for camera image input, estimation or reconstruction of 3D data is required for annotation.

In Fig. 6, we have illustrated a deep learning-based end-to-end (E2E) data annotation model. However, to train such a E2E model we have to feed with a lot of labeling data. To alleviate the data request, we propose a semi-traditional annotation framework, a hybrid with classic computer vision and deep learning, shown in Fig. 19.

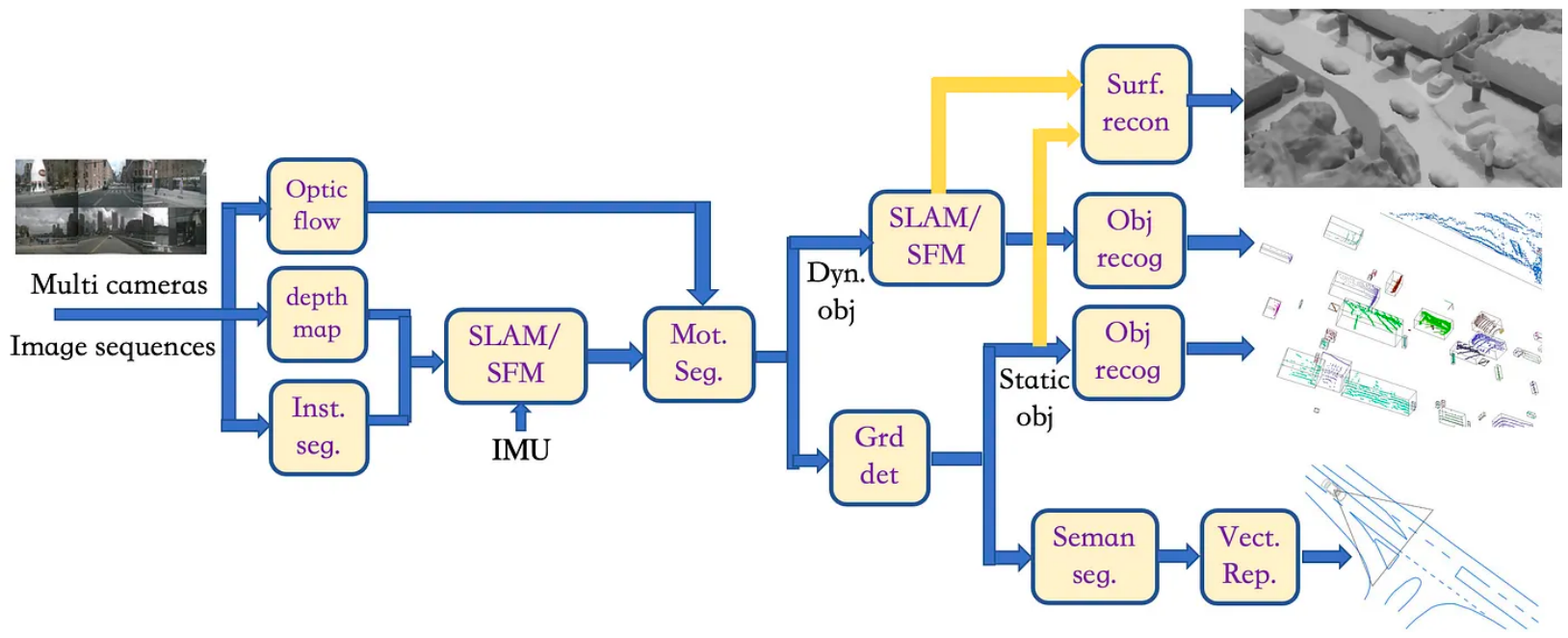

(a) Multi-cameras only

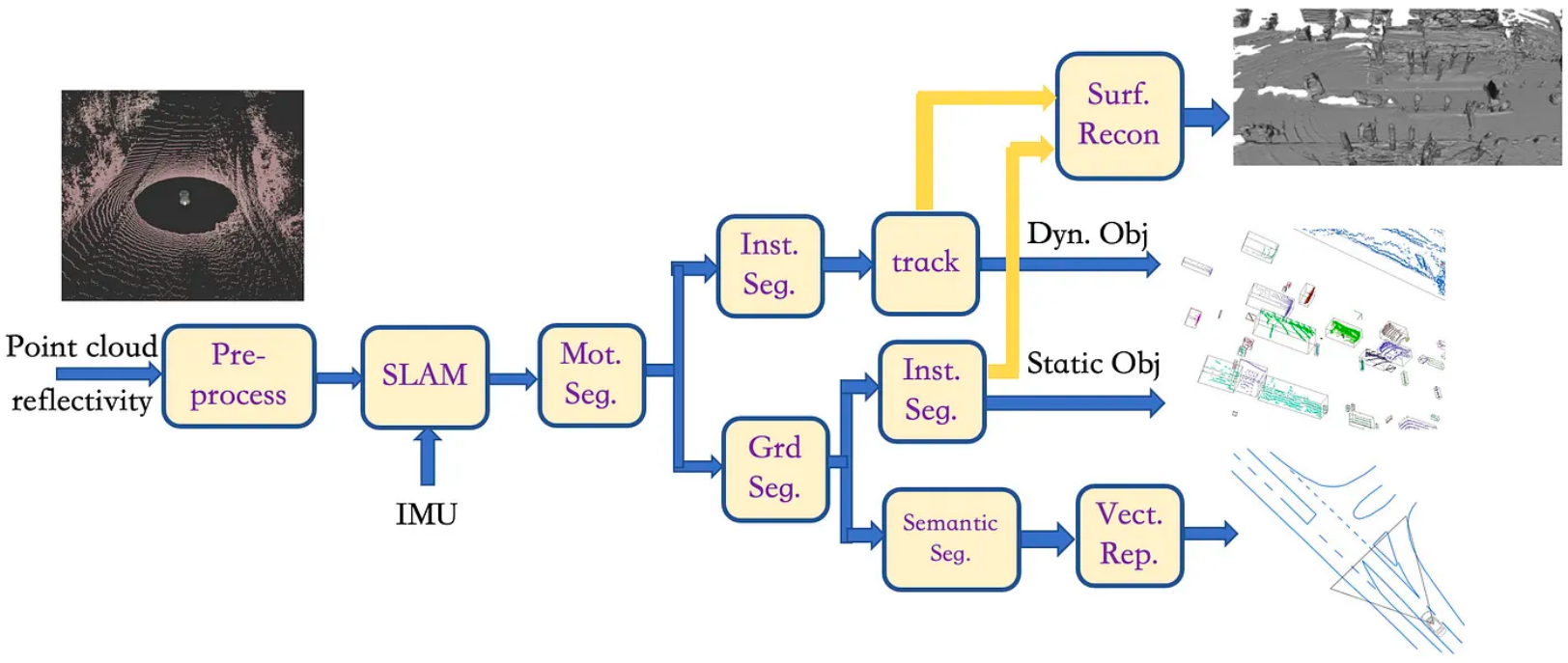

(b) LiDAR only

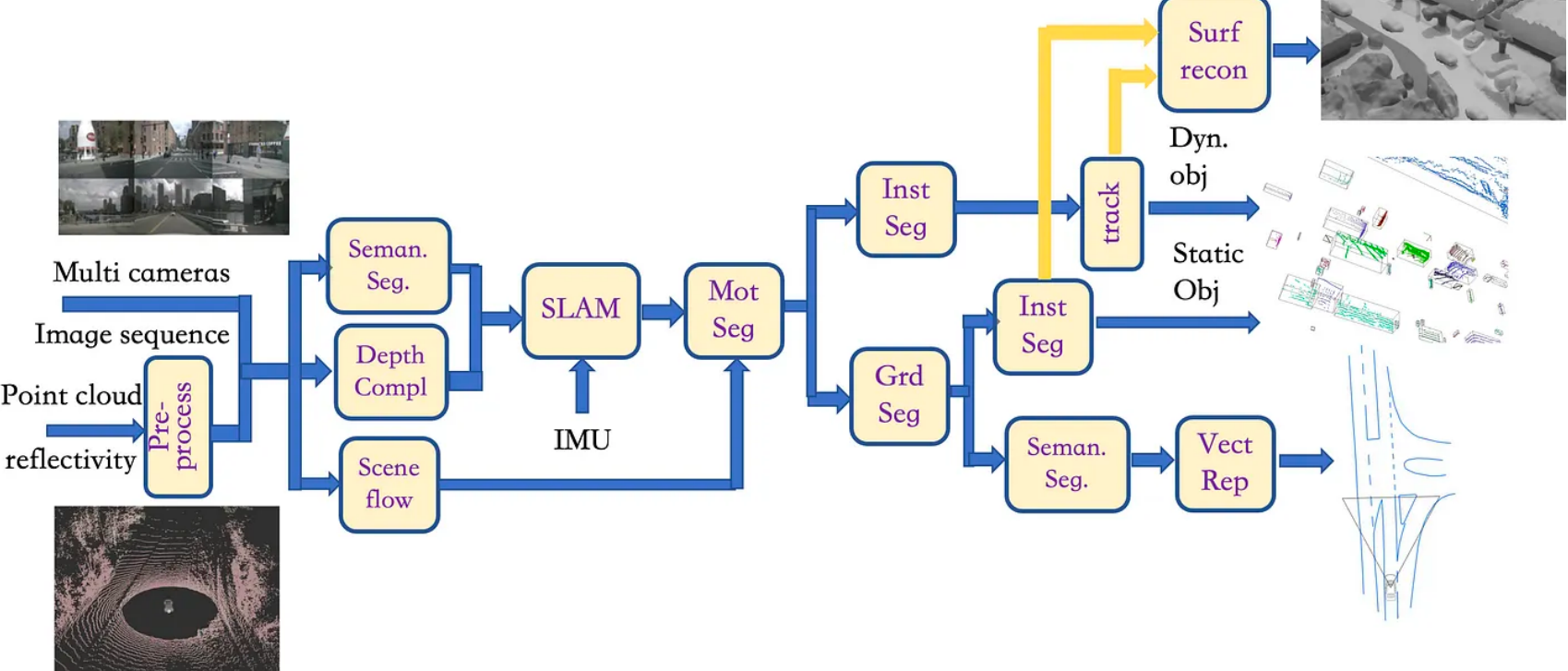

(c) Both LiDAR and multi-cameras

Fig. 19 A semi-traditional data labeling framework

For multi-cameras input only, shown in Fig. 19(a), three modules are firstly used in the image sequence of multiple cameras, namely "**inst seg**", "**depth map**", and "**optical flow**", to calculate the instance segmentation map, depth map, and optical flow map respectively; the "**inst seg**" module uses the depth learning model to locate and classify some object pixels, such as vehicles and pedestrians, the "**depth map**" module uses the depth learning model to estimate the pixel-wise motion of the two consecutive frames based on monocular video to form virtual stereo vision to infer the depth map, and the "**optical flow**" module uses the depth learning model to directly infer the pixel motion of the two consecutive frames; the three modules try to use the neural network model with unsupervised learning; based on depth map estimation, the "**SLAM/SFM**" module can obtain dense 3D reconstructed point clouds similar to RGB-D+IMU sensors; at the same time, the instance segmentation results actually allow for the removal of obstacles from the image, such as vehicles and pedestrians, while further distinguishing static and dynamic obstacles in the "**mot seg**" module based on ego-vehicle odometry and optical flow estimation; the various dynamic obstacles obtained by the instance segmentation will be reconstructed in the next "**SLAM/SFM**" module (where IMU is not input), which is similar to the SLAM architecture of RGB-D sensors and can be an extension of monocular SLAM; then, it transfers the results of "**inst seg**" to the "**obj recog**" module and annotate the 3D bounding box of the object point cloud; for static background, the "**grd det**" module will distinguish between static objects and road point clouds, so that static obstacles (such as parking vehicles and traffic cone) will transfer the results of "**inst seg**" module to the "**obj recog**" module, which annotates the 3-D bounding box of point clouds; the dynamic object point cloud obtained from the "**SLAM/SFM**" module and the static object point cloud obtained from “**grd det**” module enter the “**Surf Recon**” module to run the Poisson reconstruction algorithm; the road surface point cloud only provides a fitted 3D road surface; from the image domain “**inst seg**” module, the road surface area can be obtained; based on ego odometry, image stitching can be performed; after running the “**seman seg**” module on the stitched road surface image, lane markings, zebra crossings, and road boundaries can be obtained; then, the “**vectrep**” module is used for polyine labeling; finally, all annotations are projected onto the vehicle coordinate system, as a single frame, to obtain the final annotation.

For LiDAR input only, shown in Fig. 19(b), we undergo a “**pre-process**” module, a “**SLAM**” module, and a “**mot seg**” module; then, in the “**inst seg**” module, point cloud based detection is directly performed on moving objects those are different from the background; neural network models are used to extract feature maps from the point cloud (such as PointNet and PointPillar); afterwards, in the “**Track**” module, we perform temporal association on each segmented object to obtain annotations as dynamic objects’ 3-d bounding boxes; for static backgrounds, after the “**Grd Seg**” module, the point cloud that is judged as non road surface enters another “**Inst Seg**” module for object detection, and the annotations of the static objects’ 3-D bounding boxes are obtained; for the point cloud of the road surface, the “**Semantic Seg**” module is applied, and based on a deep learning model, we use reflection intensity to perform pixel by pixel classification of semantic objects similar to image data, that is, lane markings, zebra crossings, and road areas; the road curbs are obtained by detecting road boundaries, and finally the polyline-based annotations are made in the “**Vect Rep**” module; the tracked dynamic object point cloud and the static object point cloud obtained by the instance segmentation enter the “**surf recon**” module and run the Poisson reconstruction algorithm; finally, all annotations are projected onto the vehicle coordinate system, as a single frame, to obtain the final annotation.

For both LiDAR and multi-cameras input, shown in Fig. 19(c), we replace the “optical flow ” module in Fig. 19(a) with the “**scene flow**” module, which estimates the motion of the 3D point cloud based on a deep learning model; we also replace the “depth map” module with the “**depth compl**” module, which uses a neural network model for completion of depth which is obtained from projection of the point cloud (interpolation and “hole filling”) onto the image plane, and then inversely project it back to the 3D space to generate the point cloud; meanwhile, we replace the “inst seg” module with the “**seman. seg.**” module, which uses a deep learning model to label point clouds according to object categories; afterwards, the dense point cloud and IMU data will enter the “**SLAM**” module to estimate the odometry, and point clouds marked as obstacles (vehicles and pedestrians) will be chosen; at the same time, the estimated scene flow will also enter the "**mot seg**" module, further distinguishing between moving obstacles and static obstacles; once the moving objects pass through the "**inst seg**" module and the "**track**" module, the annotations of the moving objects are obtained; similarly, after passing through the "**grd seg**" module, static obstacles are labeled by the "**inst seg**" module; Map elements, such as lane markings, zebra crossings, and road edges, are obtained by running the "**seman. seg.**" module in the stitched road surface image and the aligned point clouds, and then enter the "**vect rep**" module for polyline labeling; the dynamic object point cloud obtained from tracking and the static object point cloud obtained from instance segmentation enter the "**surf recon**" module which runs the Poisson reconstruction algorithm; finally, all annotations are projected onto the vehicle coordinate system, as a single frame, to obtain the final annotation.

This semi-traditional annotation method is also named as *4-D annotation*, explored by Tesla's Autopilot team first. Thus, the proposed data annotation framework runs in two stages: semi-traditional 4-D annotation first and then deep learning-based end-to-end annotation.

### 12.3. *Active Learning*

Based on the detection methods of corner cases, OOD or anomaly data, the training platform of the autonomous driving machine learning model can adopt reasonable methods to absorb these incremental data. Among them, *active learning* is the most common method, which can efficiently use these valuable data. Active learning is an iterative process in which a model is learned at each iteration and a set of points is chosen to be labelled from a pool of unlabeled points using some heuristics. One of heuristics is *uncertainty estimate*, popularly used in autonomous driving field. There are two major types of uncertainty: epistemic and aleatoric uncertainty. *Epistemic uncertainty* is commonly referred to as model uncertainty, and its estimation methods mainly include the Ensemble method and the Monte Carlo dropout method; *aleatoric uncertainty* is referred to as data uncertainty, and the mostly used estimation method is probabilistic machine learning based on Bayesian theory.

**Note**: Though mostly people apply supervised learning to train the model in the data closed loop, some new techniques in machine learning like semi-supervised learning (using both labeled and unlabeled data) and even self-supervised learning (such as the popularly employed contrastive learning without labeling data) are introduced to improve the generalization, scalability and efficiency.

## 13. Conclusion

In this overview of autonomous driving, we outlined some of the key innovations as well as unsolved problems. Several deep learning-based architecture models had been proposed, i.e. BEV/occupancy perception, collaborative sensing in V2X, end-to-end autonomous driving with BEV-based perception, prediction and planning (BP3). An novel viewpoint of this paper is that we paid more attention to the data closed loop in R&D of autonomous driving. Especially, the corresponding data selection/screening and data annotation/labeling mechanisms to impulse the data closed loop, were proposed.

### *13.1 ChatGPT and SOTA Foundation Models*

At last, we want to discuss briefly the influence of *Foundation Models* to autonomous driving field and its data close-loop paradigm as well.

Recently powered by large language models (LLMs), chat systems, such as chatGPT and PaLM, emerge and rapidly become a promising direction to achieve artificial general intelligence (AGI) in natural language processing (NLP) [42]. Indeed, key creations such as large-scale *pre-training* that captures knowledge across the entire world wide web, *instruction fine-tuning*, *prompt learning*, *in-context learning*, *Chain of Thoughts* (COT) and *Reinforcement Learning from Human Feedback* (RLHF) have played significant roles in enhancing LLMs' adaptability and performance. Meanwhile, there have brought about some concerns regarding reinforcing biases, privacy violation, harmful hallucination (untruthful nonsense) and significant computer power consumption etc.

The concept of Foundation Models has been extended from NLP to other domains, like computer vision and robotics. At the same time, multi-modal input or output is realized to make broader applications. Vision-Language Models (VLMs) learn rich vision-language correlation from web-scale image-text pairs and enable zero-shot predictions on various computer vision tasks with a single VLM, such as CLIP and PaLM-E. ImageBind, an approach to learn a joint embedding across six different modalities - images, text, audio, depth, thermal, and IMU data, is proposed by Meta[43]. It actually leverages the large scale visual language models and extends zero-shot capabilities to a new modality by using its pairing with images.

The overwhelming success of Diffusion Models[50] starts from image synthesis but extends to other modalities, like video, audio, text, graph and 3-D model etc. As a new branch of multi-view reconstruction, Neural rendering, like NeRF [47] and Gaussian splatting [52], provides implicit representation of 3D information. Marriage of diffusion models and NeRF/Gaussian splatting has achieved remarkable results in text-to-3D synthesis.

### *13.2Applications of Foundation Models to ADS*

In summary, the occurrence of LLMs has made chain reactions of AGI from NLP to various domains, especially computer vision. The autonomous driving systems (ADS) will definitely be influenced by this trend. Given the enough huge data and vision language models, accompanied by NeRF/Gaussian splatting and diffusion models, the idea and operation of foundation models will generate revolutions in autonomous driving. The "long tailed" problem will be largely alleviated and the data closed loop may change to another cycle mode, i.e. pretraining + fine-tuning + reinforcement learning, let alone the ease of building the simulation platform and auto-labeling of training data for lightweight on-vehicle models.

There comes a natural thinking that we could employ abilities from LLMs to reformulate autonomous driving[54]. By combining LLM with foundation models, it is possible to utilize the human knowledge, commonsense and reasoning to rebuild autonomous driving systems from the current long-tailed AI dilemma.

NavGPT[51], a purely LLM-based instruction following navigation agent is proposed, to reveal the reasoning capability of GPT models in embodied scenes by performing zero-shot sequential action prediction for vision-and-language navigation (VLN). NavGPT can explicitly perform high-level planning for navigation, including decomposing instruction into sub-goal, integrating commonsense knowledge relevant to navigation task resolution, identifying landmarks from observed scenes, tracking navigation progress, and adapting to exceptions with plan adjustment.

Reference [53] proposed a new paradigm for autonomous driving that uses a large language model (LLM) as a cognitive agent to integrate human-like intelligence into the autonomous driving system. The approach, called Agent Driver, introduces a general tool library accessible through function calls, a cognitive memory for decision-making that includes common sense and experience knowledge, and an inference engine capable of chain-of-thought (CoT) reasoning, task planning, motion planning, and self-reflection.